\documentclass{article}

\usepackage{arxiv}

\pdfoutput=1
\usepackage[utf8]{inputenc} 
\usepackage[T1]{fontenc}    
\usepackage{hyperref}       
\usepackage{url}            
\usepackage{booktabs}       
\usepackage{amsfonts}       
\usepackage{nicefrac}       
\usepackage{microtype}      
\usepackage{amsmath}
\usepackage{cleveref}       
\usepackage{lipsum}         
\usepackage{graphicx}
\usepackage{doi}
\usepackage{rotfloat}
\usepackage{subfigure}
\usepackage{caption}
\usepackage[title]{appendix}

\newcommand*{\BeforeCaptionVSpace}{1ex}

\usepackage[linesnumbered,ruled,vlined]{algorithm2e}
\SetAlgorithmName{Algorithm}{Algorithm}{Algorithm}
\newcommand{\AlgoFontSize}{\normalsize} 
\IncMargin{0.5em}
\SetCommentSty{textnormal}
\SetNlSty{}{}{:}
\SetAlgoNlRelativeSize{0}
\SetKwInput{KwGlobal}{Global}
\SetKwInput{KwPrecondition}{Precondition}
\SetKwProg{Proc}{Procedure}{:}{}
\SetKwProg{Func}{Function}{:}{}
\SetKw{And}{and}
\SetKw{Or}{or}
\SetKw{To}{to}
\SetKw{DownTo}{downto}
\SetKw{Break}{break}
\SetKw{Continue}{continue}
\SetKw{SuchThat}{\textit{s.t.}}
\SetKw{WithRespectTo}{\textit{wrt}}
\SetKw{Iff}{\textit{iff.}}
\SetKw{MaxOf}{\textit{max of}}
\SetKw{MinOf}{\textit{min of}}
\SetKwBlock{Match}{match}{}{}

\title{PoseAction: Action Recognition for Patients in the Ward using Deep Learning Approaches}

\date{October 5, 2023}

\newif\ifuniqueAffiliation

\ifuniqueAffiliation 
\author{ \href{https://orcid.org/0000-0001-7992-9995}{\includegraphics[scale=0.06]{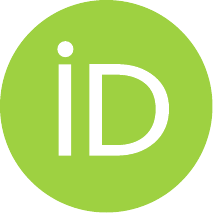}\hspace{1mm}Zherui Li}\thanks{This author is also affiliated with the Department of 
        Biomedical Engineering at Carnegie Mellon University, 5000 Forbes Ave, Pittsburgh, PA 15213.} \\
	Department of Biomedical Engineering\\
	National University of Singapore\\
	15 Kent Ridge Cres, Singapore 117583\\
	\texttt{zherui\_li@u.nus.edu} \\
	\And
	\href{https://orcid.org/0000-0002-6210-4548}{\includegraphics[scale=0.06]{orcid.pdf}\hspace{1mm}Raye Chen-Hua Yeow}\thanks{Corresponding author.} \\
	Department of Biomedical Engineering\\
	National University of Singapore\\
	15 Kent Ridge Cres, Singapore 117583\\
	\texttt{rayeow@nus.edu.sg} \\
}
\else
\usepackage{authblk}

\newbox{\orcid}\sbox{\orcid}{\includegraphics[scale=0.06]{orcid.pdf}} 
\author[1,2]{%
	\href{https://orcid.org/0000-0001-7992-9995}{\usebox{\orcid}\hspace{1mm}Zherui Li}%
}
\author[1]{%
	\href{https://orcid.org/0000-0002-6210-4548}{\usebox{\orcid}\hspace{1mm}Raye Chen-Hua Yeow\thanks{Correspondence: \texttt{rayeow@nus.edu.sg}}}%
}
\affil[1]{Department of Biomedical Engineering, National University of Singapore, Singapore 117583}
\affil[2]{Department of Biomedical Engineering, Carnegie Mellon University, Pittsburgh, PA 15213}
\fi

\renewcommand{\headeright}{}

\hypersetup{
pdftitle={PoseAction: Action Recognition for Patients in the Ward using Deep Learning Approaches},
pdfsubject={},
pdfauthor={Zherui Li, Raye Yeow},
pdfkeywords={Action recognition, Deep Learning, Computer Vision, Ward care},
}

\begin{document}
\maketitle

\begin{abstract}
    Real-time intelligent detection and prediction of subjects' behavior particularly their movements or actions is critical in the ward. This approach offers the advantage of reducing in-hospital care costs and improving the efficiency of healthcare workers, which is especially true for scenarios at night or during peak admission periods. Therefore, in this work, we propose using computer vision (CV) and deep learning (DL) methods for detecting subjects and recognizing their actions. We utilize OpenPose as an accurate subject detector for recognizing the positions of human subjects in the video stream. Additionally, we employ AlphAction's Asynchronous Interaction Aggregation (AIA) network to predict the actions of detected subjects. This integrated model, referred to as \textit{PoseAction}, is proposed. At the same time, the proposed model is further trained to predict 12 common actions in ward areas, such as \texttt{staggering}, \texttt{chest pain}, and \texttt{falling down}, using medical-related video clips from the NTU RGB+D and NTU RGB+D 120 datasets. The results demonstrate that PoseAction achieves the highest classification mAP of 98.72\% (IoU@0.5). Additionally, this study develops an online real-time mode for action recognition, which strongly supports the clinical translation of PoseAction. Furthermore, using OpenPose's function for recognizing face key points, we also implement face blurring, which is a practical solution to address the privacy protection concerns of patients and healthcare workers. Nevertheless, the training data for PoseAction is currently limited, particularly in terms of label diversity. Consequently, the subsequent step involves utilizing a more diverse dataset (including general actions) to train the model's parameters for improved generalization.
\end{abstract}

\keywords{Action recognition\and Deep learning\and Computer vision\and Ward care}

\section{Introduction}
\label{ch:intro}

Monitoring the vital signs of the patients in the ward is very important and this can easily be achieved by IoT (Internet of Things) bedside monitors or other equipment. Equally important is the real-time intelligent recognition of the patient's behavior (especially the movements, or say, actions) in the ward. This has the advantage of reducing in-hospital care costs and increasing the efficiency of healthcare workers, which is especially true for scenarios where there are fewer staff on duty in the ward at night or during peak admission periods. In particular, such a method can be used to monitor the behavior of some subjects who need to take medication regularly or get up and do exercise. Also, based on the subject's behavior, other methods can be used to help determine the progress of the patient's disease and even his or her mental state during recovery. Thus, developing a model to predict patients' actions becomes important for modern hospital wards that are transforming into smart healthcare. Therefore, in this work, we first proposed the use of computer vision (CV) and deep learning (DL) methods for human subject detection and action pattern recognition in the ward area.

\subsection{From Traditional Healthcare to Smart Healthcare for Wards}
In modern hospitals, patients from different departments are placed in different ward areas for treatment or rehabilitation. These patients often suffer from similar diseases or require similar treatment. For example, a patient in a hematology ward area may have a hematologic-related disease such as leukemia, while a patient in a neurosurgery ward may have a cranial injury that requires or has already undergone surgery. Patients are usually admitted and housed in the ward at various stages of their disease. What they all have in common is the need for family or medical staff to care for them and to respond to their condition and disease progression.

In traditional wards, patients are routinely evaluated by healthcare professionals during hospitalization by measuring vital signs. These measurements are critical to preventing patient deterioration, reducing morbidity and mortality, shortening the length of stay, and reducing patients' costs~\cite{mok2015attitudes}. Although there are differences in the vital signs of interest in different hospitals around the world, common vital signs measured by healthcare professionals include heart rate (HR), oxygen saturation (SpO2), and blood pressure (BP)~\cite{elliott2012critical}. At the same time, the healthcare professionals will perform further tests and examinations on the possible causes of the patient's disease to assess the patient's status in more detail and what corrective measures may be needed. This can involve significant labor costs. And this can also be fatal in areas where healthcare resources are relatively scarce, or in the early stages of a pandemic when there is a peak period of admissions, which can lead to situations such as the early stages of a new COVID-19 pandemic where new admissions cannot be treated in a timely manner because healthcare professionals are also infected with the virus. This may even result in unnecessary deaths of healthcare professionals~\cite{bandyopadhyay2020infection}. Therefore, for the foreseeable future, traditional wards will no longer be able to meet the needs of patients for quality care in countries and regions with a high level of aging.

With the growth of the IoT and Internet of Health Things (IoHT) technologies and the introduction of the Smart Ward concept, many patients' vital signs can be collected and recorded in a semi-automated or fully automated manner, thus reducing the pressure on healthcare professionals to care for patients in the ward~\cite{da2018internet}. However, the information that can be collected in these automated ways is currently very limited, and the assessment of pain indices and levels of awareness, for example, still requires subjective measurements by healthcare professionals at this stage~\cite{jones1979glasgow,fink2000pain}. The good news is that, to some extent, this information can be estimated through the behavioral characteristics of the patient. Also, patient behavioral data can be used as a basis for preventing unintentional injuries in the ward area. For example, a patient may unconsciously get up from the bed and fall, which may cause additional harm to the patient if not detected by the nurse on duty. Therefore, we believe that in addition to intelligent monitoring of patients' vital signs, smart wards also need to monitor patients' behavioral data. This has the advantage of reducing the pressure on healthcare professionals and increasing the number of patients that can be seen by each staff member while keeping the number of staff and the quality of care the same.

\subsection{Introducing Computer Vision and Deep Learning Methods to Monitor Patients' Actions}
\subsubsection{AlphAction - Multiple-Person Action Recognition}
Tang \textit{et al.} at SJTU (Shanghai Jiao Tong University) innovatively proposed the AlphAction model trained on the AVA dataset (a video dataset of atomic visual action)~\cite{gu2018ava}, which includes a YOLOv3-based detector for subject detection and an Asynchronous Interaction Aggregation (AIA) detector for predicting action patterns~\cite{tang2020asynchronous}. The core architecture of the AlphAction model will be further discussed in \autoref{ch:review-3-1}. By using the model parameters given by the AlphAction project, which includes the top 60 most common classes of actions in the AVA dataset, such as $\texttt{stand}$, $\texttt{talk to}$, $\texttt{sit}$, $\texttt{answer phone}$, and $\texttt{walk}$, the subject action recognition performance of the AlphAction (i.e., the \textit{Ours} class) is shown in \autoref{fig:intro-1} below. 

\begin{sidewaysfigure}[htp]
\centerline{\includegraphics[width=23cm]{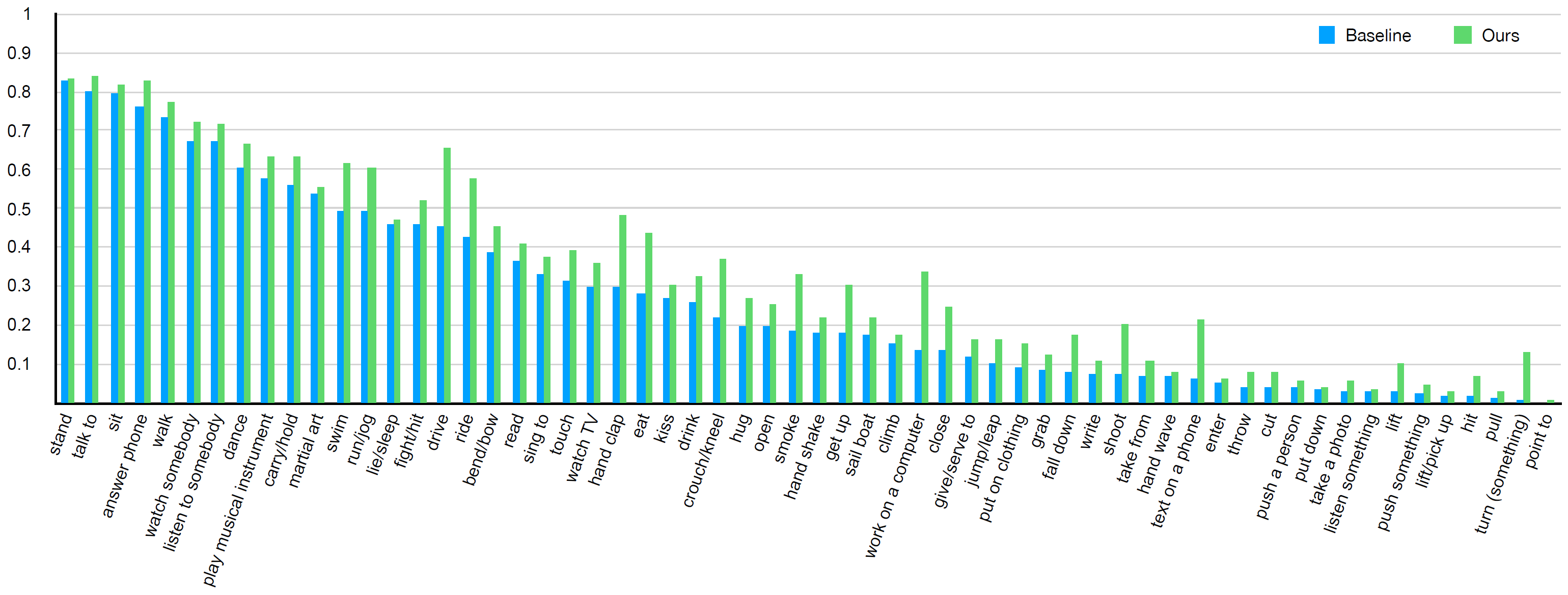}} 
\caption{Tested action recognition classes and performance of the AlphAction model and the baseline model on the validation set of the AVA v2.2 dataset. Figure reproduced from Tang \textit{et al.} (2020), \textit{ECCV}.}
\label{fig:intro-1}
\end{sidewaysfigure}

However, it is not realistic to predict the actions of patients within a ward area directly using AlphAction. The reason for this comes from three main aspects:
\begin{itemize}
    \item[(1)] \textbf{The categories of actions included in the AVA dataset do not satisfy the recognition of the categories of actions that are commonly seen and risky in the ward.} As can be seen from \autoref{fig:intro-1}, only a few action classes such as $\texttt{fall down}$ are of particular concern in the ward. Therefore, when we further train the AIA detector for PoseAction, we need to define the common risky action classes in the ward that need to be paid attention to, and thus find a suitable video dataset for training.\\
    
    \item[(2)] \textbf{The YOLOv3 model is not accurate enough to detect common patient subjects in the ward area.} Pre-experiments revealed that the human subject detector based on the YOLOv3 model used by AlphAction does not detect bedridden and covered patients well. As a result, the AIA detector does not predict the actions of these subjects. This reduces the reliability of AlphAction for direct application to ward scenarios.

    \item[(3)] \textbf{The performance by using the existing AIA network parameters is not good enough.} As can be seen in \autoref{fig:intro-1}, AlphAction has a recognition accuracy of more than 50\% for less than 1/3 of the total number of categories on the AVA v2.2 dataset used by the authors. This is certainly unacceptable for a patient action recognition model that will be applied to medical-related scenarios. Therefore, it is necessary to improve the recognition accuracy from the perspective of the video quality of the training dataset.
\end{itemize}

\subsubsection{OpenPose - Multi-Person 2D Pose Estimation}
Due to the lack of accuracy of YOLOv3's detection of human subjects in wards, we replaced the YOLOv3 detector with an OpenPose model-based detector. The OpenPose model was proposed by Cao \textit{et al.} at CMU (Carnegie Mellon University)~\cite{cao2017realtime,simon2017hand,wei2016convolutional}. OpenPose was initially used for the real-time generation of human subject key points and for 2D pose estimation. The core architecture of the OpenPose model will be further discussed in \autoref{ch:review-2-2}. We abstracted the OpenPose model and implemented a keypoint-based detector to obtain the smallest rectangular bounding box (\textit{bbox}) of human subjects. This is consistent with the role of the YOLOv3 detector in the original AlphAction model but with better accuracy. Through the pre-experiment, we also found that after replacing the OpenPose as the detector, the patient lying under the quilt, which could not be detected before, could be well detected. This provides the basis for further building and training the PoseAction model.

\subsubsection{PoseAction - An Integrated Model for Better Human Detection and Action Recognition}
In summary, PoseAction is an integrated model that inherits the advantages of both the AIA network of the AlphAction model and the OpenPose model. Its main contributions are the following three points:
\begin{itemize}
    \item[(1)] \textbf{Human subject detection and action recognition in complex scenarios.} For patients with partially obscured torsos, PoseAction can better identify these subjects and predict their action patterns. This allows healthcare professionals to intervene in a timely manner to reduce the likelihood of accidental injury in the ward area.
    
    \item[(2)] \textbf{Provide optimized action labels that need attention in the ward.} In this work, video clips of 12 common medical-related scenarios from the NTU RGB+D and NTU RGB+D 120 datasets were used to train the parameters of the PoseAction AIA detector. Such scenarios include $\texttt{staggering}$, $\texttt{falling down}$, $\texttt{headache}$ and $\texttt{chest pain}$ as labeled~\cite{shahroudy2016ntu,liu2020ntu}. Using these optimized labels can better determine the abnormal actions of patients in the ward area.

    \item[(3)] \textbf{Able to post-generate with existing video clips, as well as process real-time video streams.} Since patient care in the ward emphasizes real-time, PoseAction's real-time processing mode allows for the processing of video streams from surveillance cameras, for example. This has the advantage of reducing the time window between the onset of a patient's abnormal action and its detection by healthcare professionals.
\end{itemize}

\subsection{Synopsis}

\begin{figure}[htb]     
\centerline{\includegraphics[width=\columnwidth]{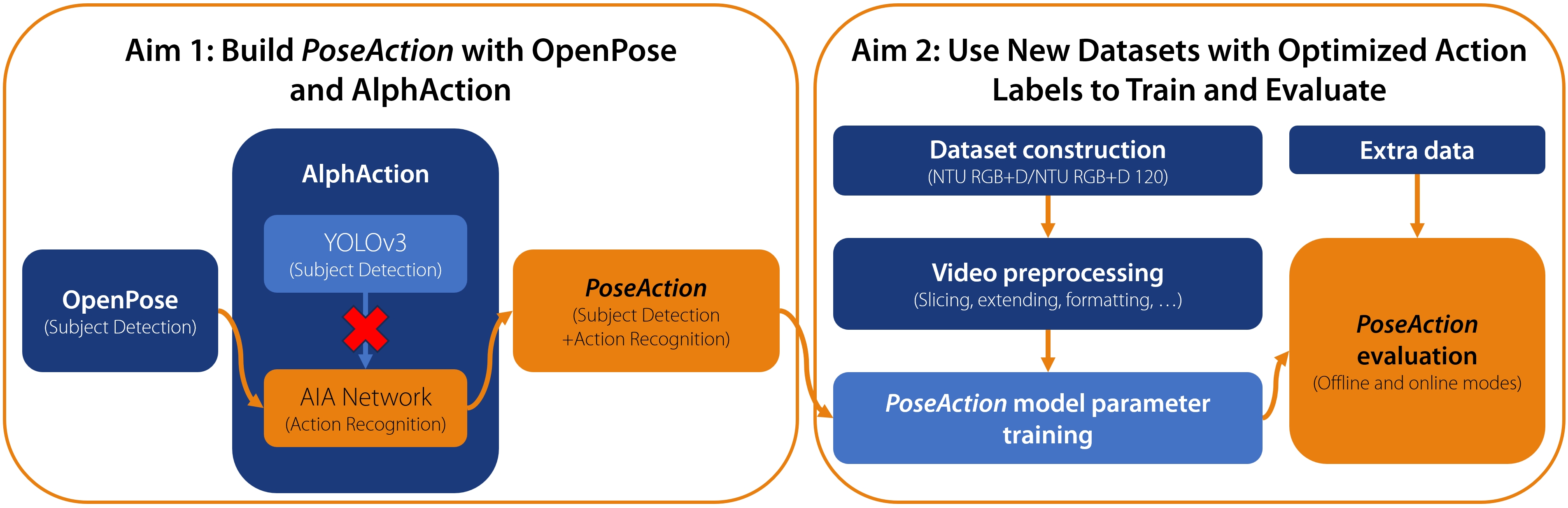}}
\vspace{\BeforeCaptionVSpace}
\caption{Aims and realization of the PoseAction.} 
\label{fig:intro-2} 
\end{figure}

This work is focused on developing the proposed PoseAction model, a better deep-learning solution for patient subject detection and action recognition in ward areas, and AIA network parameter training with preferred medical-related labels. The primary contribution of this study is to present the proposed model that offers enhanced accuracy in identifying common hazardous actions within a hospital ward environment. This study addresses the existing gap in action recognition models specifically tailored for ward scenarios. Additionally, it overcomes the challenge of low detection accuracy when directly applying existing action recognition models in this domain. The main aims of this work and the flowchart of their realization are shown in \autoref{fig:intro-2}.

The rest of this paper is organized as follows:

In \autoref{ch:review}, we conduct a literature review. Here we present in detail the considerations for ward design in modern hospitals, the need for smart wards, and the possible and preventable measures for accidental injuries in wards. Secondly, the principles and development of common human subject detection methods are briefly introduced, and the basic structure of human posture estimation methods and OpenPose models are further extended. Finally, the difference between human action recognition and prediction and the development of human behavior recognition are briefly introduced, and then the basic structure and implementation of the AlphAction model are introduced.

In \autoref{ch:methods}, we provide the methods of this work. We present the methods for human subject detection using OpenPose instead of YOLOv3 by calculating the smallest rectangular bounding box (\textit{bbox}) from key points. Then, we discuss the overall architecture of the proposed PoseAction model by presenting its two modes. Lastly, we introduce the training and validation dataset formation methods using the NTU RGB+D/NTU RGB+D 120 datasets, as well as the key training parameters and the performance evaluation metrics.

In \autoref{ch:results}, we provide the results of this work. We show the loss and accuracy curves based on the NTU RGB+D/NTU RGB+D 120 dataset in training, as well as the inference AP and mAP curves for every 10,000 iterations in training based on the validation set. Also, we use the final obtained model weights for validation to obtain specific performance metrics for PoseAction under different configurations. In addition, we also show the results of tests performed on additional videos to demonstrate the generalization of the PoseAction.

In, \autoref{ch:discussion}, we analyze the performance of the PoseAction and try to explain the reasons that lead to its excellent performance. In addition, we discuss some other possible application scenarios of the PoseAction, such as face-blurring features for patient privacy protection. Finally, we also discuss the limitations of the current version of the PoseAction, and the reasons why we cannot directly share our generated training and validation sets.

Finally, in \autoref{ch:concl}, we briefly summarize the objectives, significance, innovations, and results of this work. Furthermore, some feasible future research directions based on the PoseAction are also proposed.

\section{Literature Review}
\label{ch:review}

\subsection{Ward Design in Modern Hospitals and Accidental Injuries in Wards}
\label{ch:review-1}
The design of modern hospital wards varies from country to country and from hospital to hospital. However, in general, the design of hospital rooms is usually decided by architects and medical professionals, with little direct consultation and preference from patients~\cite{pattison1996effect}. The result of this may not only lead to less efficient nursing care within the ward, but also to adverse psychological and physical factors affecting the patient. In a review of hospital ward layouts and nurse staffing, Seelye outlines aspects of ward design and operation, such as patient privacy, nursing efficiency, patient dependency, and ward organization, and further describes methods for establishing staffing levels that take ward design into account. It concludes that effective and efficient nursing-related ward layouts should consider minimizing movement distances and facilitating maximum contact between healthcare professionals and patients~\cite{seelye1982hospital}. Also, Pattison \textit{et al.} noted that different ward layouts can positively or negatively affect the psychological factors of postoperative patients. For example, patients recovering in the Bay Ward had better sleep quality and mental health, while the traditional Nightingale Ward showed the opposite impact factors on these indicators~\cite{pattison1996effect}. In conclusion, it is clear from previous studies that the more beneficial ward design for patients tends to be one with individual cubicles and adequate staff to respond to the patient's medical progress and psychological needs in a timely manner. However, in many countries and regions where aging is increasing or where medical resources are relatively scarce, such a ward design is difficult to achieve.

To alleviate this contradiction, the concept of Smart Wards was developed. In general, hospital wards are relatively "mobile", especially during certain unpredictable peak admission periods. As mentioned earlier, hospitals may often face a shortage of medical and nursing staff and a low doctor-patient ratio. As a result, routine vital sign monitoring may be delayed or untimely, leading to overnight deterioration of the patient's condition. However, monitoring patients' vital signs is a labor-intensive task for traditional wards, requiring a large number of healthcare professionals to be on duty throughout the day~\cite{clifton2015errors}. As a result, smart wards will become commonplace in the foreseeable future. In a smart ward, patients' vital signs can be monitored by a number of wireless sensors, thus allowing early prediction and reporting of the progress of some diseases for timely intervention by medical staff~\cite{breteler2018reliability}. At the same time, smart wards can be equipped with features such as real-time voice assistants to enable room temperature and lighting control, as well as paging of medical staff for patients. In addition, as Joshi \textit{et al.} suggests, artificial intelligence (AI) in smart wards will be used to eliminate many labor-intensive administrative tasks, thus freeing up more time for healthcare professionals to spend with patients~\cite{joshi2018time}. Techniques based on Big Data, such as machine learning (ML), can transform caregiving strategies in the ward from a model of symptomatic treatment of current diagnoses to one of prediction and prevention of potential disease risks, and ultimately to AI-supported treatment decisions~\cite{darzi2018better}. For the present, Da Costa \textit{et al.} suggest that the IoHT can be used to enable intelligent monitoring of patient's vital signs in smart wards. IoHT belongs to a research area that uses wearables, biosensors, and other medical devices to improve patient data management in hospitals, with the ultimate goal of reducing hospital stays and improving healthcare delivery to patients~\cite{da2018internet}. Cai \textit{et al.} proposed an intelligent ward collaboration system based on brain-computer interfaces (BCI) and IoT to improve patient self-care~\cite{cai2022toward}. Hunter \textit{et al.} have developed a solid-state nitric oxide (NO) sensor for asthma monitoring, providing a more sensory-free and reliable solution for asthma patients who are hospitalized or recovering at home~\cite{hunter2011smart}. In addition, in response to the need for care of patients admitted intensively during the COVID-19 pandemic and the need to reduce the risk of infection for healthcare professionals, Yang \textit{et al.} suggested that a multi-function teleoperated robot could be used to care for patients in isolation ward rooms~\cite{yang2020keep}. Further, we can extend the concept of Smart Wards to the field of Tele-rehabilitation. For patients discharged from COVID-19 infection and in the home rehabilitation phase, physiological indicators such as HR and SpO2 can also be monitored using remote equipment based on IoT technology~\cite{sakai2020remote,andritoi2022use,pronovost2022remote}. These data can be reported to the hospital in real-time for the physician to determine the patient's recovery status and guide the medication and rehabilitation exercise of the recovering patient.

In addition to predicting and treating the diseases that lead to patient admission, another focus of care in hospital wards is to reduce the likelihood of accidental injuries. In general, healthcare-related accidents that can occur while patients are in the ward include hospital-acquired infections. Non-medical accidents include falls from heights, slips, and bumps. For hospital-acquired infections, Ellison \textit{et al.} showed in a clinical trial that ward design did not significantly affect the risk of hospital-acquired infections or colonization for patients~\cite{ellison2014hospital}. In contrast, Stiller \textit{et al.} found that single-patient rooms were advantageous in reducing rates of healthcare-associated colonization and hospital-acquired infections, and that having hand sanitizer dispensers at the patient's bedside facilitated infection control~\cite{stiller2016relationship}. Non-medical accidents such as falls are common in children and senior patients, and these accidents have a greater impact on this group of patients~\cite{levene1991accidents}. Clearly, hospital wards also need to take steps to prevent these accidents or to intervene in a timely manner after they occur. In a previous clinical study, Drahota \textit{et al.} found that the use of shock-absorbing flooring with better "push/pull" characteristics reduced patient injury rates~\cite{drahota2013pilot}. In another study, Haines \textit{et al.} found that the introduction of low-low (i.e., low-profile) beds did not reduce the incidence of falls or fall injuries~\cite{haines2010pragmatic}. Therefore, for smart wards, one of the points that can be addressed is to reduce the delay in the intervention of medical staff after a non-medical accident, such as a fall, by monitoring the patient's movement patterns and predicting his or her actions.

\subsection{Detection Methods for Human Subjects}
\label{ch:review-2}
The problem of detecting human subjects can be formulated simply as: given an image or video sequence, locate all human subjects. This problem corresponds to determining the region, usually the smallest rectangular bounding box in the image or video sequence that encloses the humans~\cite{nguyen2016human}. At the same time, this problem has received a lot of attention in the field of CV and pattern recognition in the last decade or so, because the detection of human subjects has supported the development of many new applications, such as crowd estimation and emergency warning for public areas, and the need for pedestrian detection for autonomous vehicles. There are three main types of classical human subject detection frameworks:

\begin{itemize}
    \item[(1)] \textbf{Window-based approach:} This approach assumes that every human subject can be surrounded by a detection window. Usually, in the absence of any prior knowledge about the size and position of human subjects, windows are extracted at various scales and positions using this method. Thereafter, some overlapping windows that have been classified as human need to be merged to obtain the final result~\cite{dalal2006finding}.

    \item[(2)] \textbf{Background subtraction approach:} When the input for human subject detection is video, human detection can be performed by separating moving "objects" from the "background" by computing the difference between the current frame and the reference background in a pixel-by-pixel manner. However, background subtraction usually requires a fixed camera to obtain a stable and constant video and a reference background without human subjects~\cite{stauffer2000learning}.

    \item[(3)] \textbf{Stereo/depth information-based approach:} When the input for human detection is a stereoscopic image or video, the detection can be achieved by contour extraction of different subjects based on depth information~\cite{zhao2000stereo,xia2011human}.
\end{itemize}

\subsubsection{Common Human Detection Methods}
\label{ch:review-2-1}
Intuitively, the workflow of a human subject detector is divided into two main steps: first finding the focal regions that may be human subjects, and then determining whether human subjects are present in these regions. Here we briefly summarize the method used to determine the presence or absence of human subjects in a given region, i.e., a binary classifier (with a \textit{is human} or \textit{not human} classification result). Such a classifier can be broadly classified into two categories, i.e., generative and discriminative methods~\cite{nguyen2016human}.

The generation method aims at constructing models of human subjects, such as shape models~\cite{gavrila2007bayesian}, structure models~\cite{mikolajczyk2004human}, and appearance and structure models~\cite{barinova2012detection}. These models can then be used to match and score different regions to achieve classification.

Discriminant methods are a more common and widely developed and discussed class. This class of methods includes different implementations such as support vector machine (SVM), neural network (NN), and transfer learning (TL). Among them, SVM is usually used to classify human and non-human descriptors by maximizing the interval between the two classes~\cite{dalal2006human}. In addition, many previous works have fully discussed the lack of classification accuracy of traditional linear SVMs due to the diversity of both non-human and human subjects, and proposed improved SVM methods such as piecewise linear SVM (PL-SVM) for classification~\cite{ye2012human}. NN enables the simultaneous completion of sub-tasks of human detection, such as feature selection, subject description, and occlusion processing organized into different layers of a deep convolutional neural network (CNN), and the parameters of each layer are learned jointly by the network~\cite{ouyang2013joint}. However, it is important to note that the performance of NN-based classifiers is very dependent on the design of the network structure and the selection and construction of the training dataset. TL is a solution to the problem that trained classifiers perform well on their training and validation sets but not on other datasets for testing. This process of transferring "known knowledge" from the original dataset to the new dataset can be achieved by directly extending the original training set with a small number of labeled samples from the new dataset~\cite{cao2011rapid}.

\subsubsection{Deep Learning Based Human Pose Estimation and OpenPose}
\label{ch:review-2-2}
\begin{figure}[!t]     
\centerline{\includegraphics[width=0.9\columnwidth]{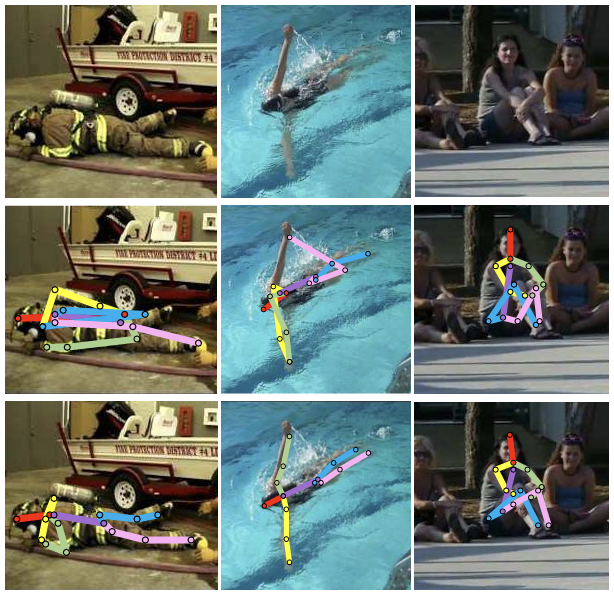}}
\vspace{\BeforeCaptionVSpace}
\caption{Example of single-person pose estimation. The first row: original images; The second row: the results of the stacked hourglass network (HG); The third row: the results of Adversarial PoseNet. Figure reproduced from Chen \textit{et al.} (2017), \textit{ICCV}.} 
\label{fig:review-1} 
\end{figure}

\begin{figure}[!t]     
\centerline{\includegraphics[width=0.9\columnwidth]{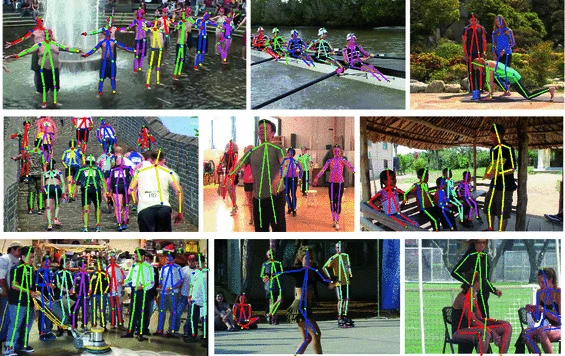}}
\vspace{\BeforeCaptionVSpace}
\caption{Example of multi-person pose estimation. Results by DeeperCut. Figure reproduced from Insafutdinov \textit{et al.} (2016), \textit{ECCV}.} 
\label{fig:review-2} 
\end{figure}

Based on human subject detection, human pose estimation methods are further developed in recent years. It aims to automatically localize human body parts (e.g., head, torso, and limbs) from images or videos~\cite{dang2019deep}. Under strict conditions, human pose estimation methods can identify and label the key points of a given human subject in an image or video (\autoref{fig:review-1})~\cite{chen2017adversarial}. Furthermore, under general conditions, human pose estimation methods require key point identification and annotation of all human subjects in an image or video, even though some of them may not be fully displayed in a single frame of a video (\autoref{fig:review-2})~\cite{insafutdinov2016deepercut}.

For the most basic 2D human pose estimation methods, there are two main types: single-person pose estimation and multi-person pose estimation. Single-person pose estimation methods are divided into two categories according to the way of predicting key points: direct regression-based methods and heat map-based methods. The former uses direct regression of the output feature maps to estimate key points, while the latter first generates a heat map and predicts key points based on the heat map. Multi-person pose estimation methods can also be divided into two categories: top-down methods and bottom-up methods. Top-down methods are broadly divided into these steps: locating human subjects in the image or video, then estimating their body parts, and finally estimating their pose. The bottom-up approach has similar steps but in reverse order: first estimate human parts in the image or video and then estimate the pose. The method jointly labels the candidate parts of the detected parts and associates the candidate parts with individuals by pairwise score regression of the spatial offset of the detected parts~\cite{dang2019deep}. In addition, some 2D or 3D human pose estimation methods using radio signals (e.g., Wi-Fi) have been proposed recently~\cite{geng2022densepose}. However, since this work used RGB image- and video-based methods, no additional discussion of these new developments is provided. 

Additionally, another scenario in the clinical field where human pose estimation needs to be implemented is the operating room (OR). For example, Özsoy \textit{et al.} point out that ORs are highly variable, unpredictable, and irregular environments with extensive interactions between different subjects, which makes it extremely challenging to model surgeries~\cite{ozsoy20224d}. Relying solely on 2D pose estimation is undoubtedly difficult in such complex scenarios that require sophisticated action recognition. Therefore, in such scenarios, not only the introduction of 3D pose estimation can be considered, but also concepts such as time-varying, scene graphs, and semantic scene graphs (SSGs) can be introduced to describe and summarize the surgical scenarios in order to expand the field of surgical data science~\cite{johnson2015image,ozsoy20224d}.

\begin{figure}[!ht]     
\centerline{\includegraphics[width=\columnwidth]{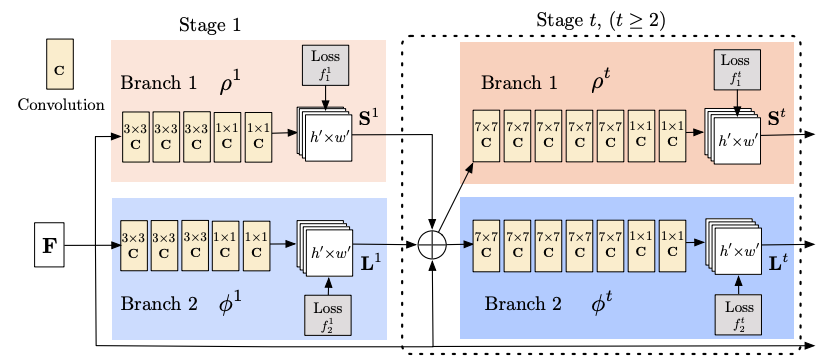}}
\vspace{\BeforeCaptionVSpace}
\caption{Core architecture of the OpenPose. Each stage in the first branch predicts Part Confidence Map (PCM) $\mathrm{\textbf{S}}^\textbf{t}$, and each stage in the second branch predicts Part Affinity Fields (PAF) $\mathrm{\textbf{L}}^\textbf{t}$. After each stage, the predictions from the two branches, along with the image features, are concatenated for the next stage. Figure reproduced from Cao \textit{et al.} (2017), \textit{CVPR}.} 
\label{fig:review-3} 
\end{figure}

The OpenPose model used in this work is a method that enables 2D and 3D pose estimation, facial expressions, and finger movements of human subjects. It is suitable for single and multi-person pose estimation with excellent robustness. OpenPose is also the first real-time multi-person pose estimation application based on DL. The concept of Part Affinity Fields (PAF) was proposed in the earlier work of its authors. This is a representation consisting of a set of flow fields that encode unstructured pairwise relationships between a variable number of body parts~\cite{cao2017realtime}. And OpenPose mainly demonstrates that refinement of the confidence map by removing body parts while increasing the network depth (PAF refinement is essential to improve accuracy) can lead to a faster and more accurate model. Also, OpenPose introduces a detector that combines body and foot key points, and demonstrates that the combination of the two detection methods not only reduces inference time compared to running them separately, but also maintains their respective accuracy. The core architecture of the OpenPose model is shown below in \autoref{fig:review-3}.

\subsection{Recognition Methods for Human Actions}
\label{ch:review-3}
\begin{figure}[htb]
    \centering
    \subfigure[Action Prediction]{
        \textsf{\includegraphics[width=0.45\columnwidth]{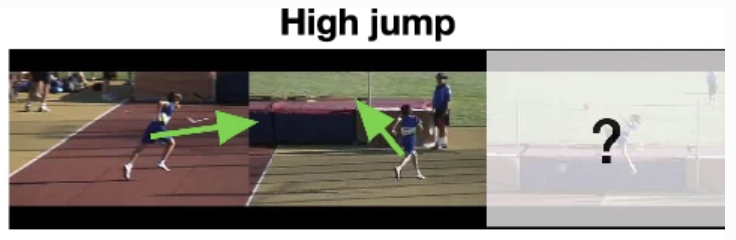}}
        \label{fig:review-4-a}
    }
    \quad
    \subfigure[Action Recognition]{
        \textsf{\includegraphics[width=0.45\columnwidth]{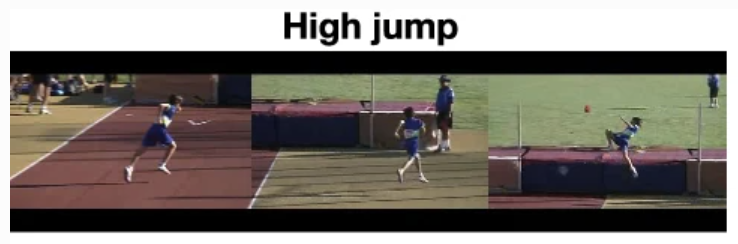}}
        \label{fig:review-4-b}
    }
    \vspace{\BeforeCaptionVSpace}
    \caption{A simple demonstration of the difference between action prediction and action recognition. Figure reproduced from Kong \textit{et al.} (2022), \textit{IJCV}.}
    \label{fig:review-4}
\end{figure}

Thanks to the rapid development of technologies such as CV and ML in the past few decades, machines are beginning to "understand" human behavior in images or videos. In general, this "understanding" can be manifested in two ways: one is the prediction of human actions, and the other is the recognition of human actions. The goal of the former is to infer labels from temporally incomplete videos (Figure \autoref{fig:review-4-a}). These labels can be action categories (i.e., early action classification) or motion trajectories (i.e., trajectory prediction). The latter objective is simpler, which is to infer action labels from videos containing complete action executions (Figure \autoref{fig:review-4-b})~\cite{kong2022human}.

Since the AlphAction model used in this work is a human action recognition model, we only focus on the development and current status of this class of models here. Human action recognition models have been popular in the past decades because of their wide range of applications, such as video retrieval, video surveillance, and human-machine interaction~\cite{bobick2001recognition,tang2012learning,ciptadi2014movement,singh2010muhavi}. At the same time, the construction of human action recognition models requires the solution of two problems: action representation and action classification, since computers cannot obtain the behavioral information of human subjects directly from images or videos. Similarly, these two problems have been well elucidated and managed to be solved from several perspectives in the past time~\cite{raptis2013poselet,carreira2017quo}. In general, a human action recognition model first converts the human subject regions detected in the video into feature vectors~\cite{wang2016robust}, and then classifies them into various action categories~\cite{liu2011recognizing}. In recent times, with the application of DL and deep neural networks (DNN) in this field, many end-to-end trainable frameworks have been proposed to implement human action recognition~\cite{tran2015learning,ji20123d}. The advantage of these new models is their integration of both action representation and action classification modules, as well as their ability to achieve better classification performance. Obviously, one factor that affects the performance of such models is the depth of the network while the other is the size of the dataset used to train the network parameters. In the former case, deep networks are currently more common in the construction of human action recognition models, but shallower models may have better results for small datasets~\cite{kong2022human}.

\subsubsection{AlphAction in Detail}
\label{ch:review-3-1}
\begin{figure}[!ht]     
\centerline{\includegraphics[width=\columnwidth]{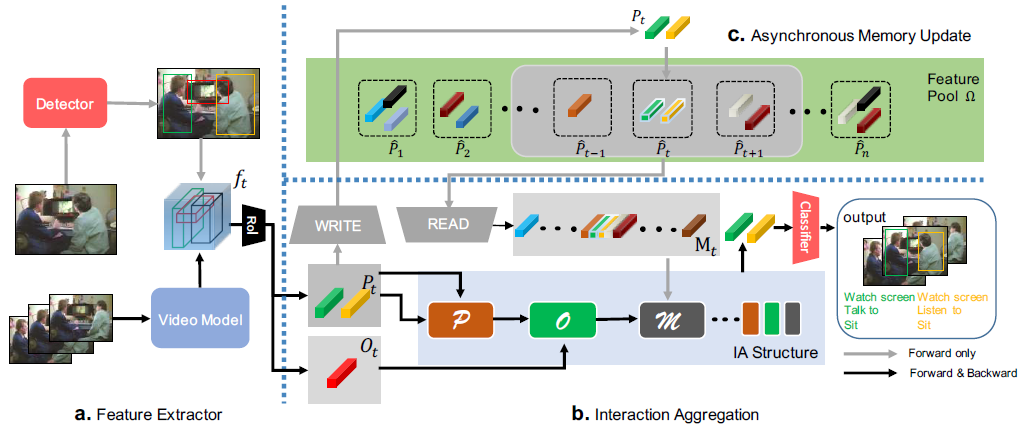}}
\vspace{\BeforeCaptionVSpace}
\caption{Core architecture of the AlphAction. Features from human subjects and objects around are first extracted. These features are then fed into the Interaction Aggregation (IA) network to integrate multiple interactions. The Asynchronous Memory Update (AMU) algorithm reads memory features from the feature pool $\Omega$ and writes fresh human features to it. Figure reproduced from Tang \textit{et al.} (2020), \textit{ECCV}.} 
\label{fig:review-5} 
\end{figure}

The AlphAction model used in this work aims to detect the actions of multiple people in a video. It is the first open-source project to achieve 30+ mAP (32.4 mAP) on an AVA dataset using a single model, where mAP stands for mean average precision, a common evaluation metric for multi-label classification tasks. It uses a SlowFast network as the baseline backbone and is built using ResNet-50 and ResNet-100 structures. The authors used the Kinetics-700 dataset to pre-train the network parameters, so the subsequent training work was performed on the basis of migration learning~\cite{tang2020asynchronous}. In their work, the authors classify interactions into three categories, namely Person-Person Interaction (e.g., conversation), Person-Object Interaction (e.g., picking up something), and Temporal Interaction (e.g., events with strong temporal correlation such as opening and closing doors). The main contributions of the AlphAction are the proposed Interaction Aggregation (IA) structure and the Asynchronous Memory Update (AMU) algorithm. The core architecture of the AlphAction model is shown below in \autoref{fig:review-5}.

\begin{figure}[!ht]     
\centerline{\includegraphics[width=0.9\columnwidth]{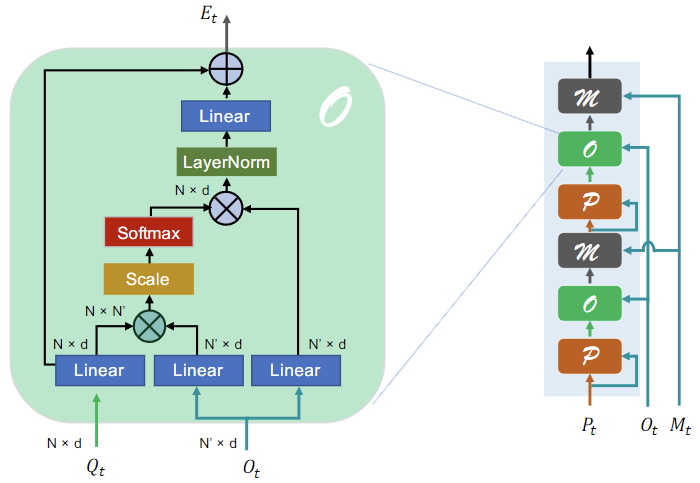}}
\vspace{\BeforeCaptionVSpace}
\caption{Left: Network architecture of the O interaction block; Right: Architecture of the serial IA structure. Figure reproduced from Tang \textit{et al.} (2020), \textit{ECCV}.} 
\label{fig:review-6} 
\end{figure}

\begin{figure}[htb]
    \centering
    \subfigure[Dense Serial IA]{
        \textsf{\includegraphics[width=0.3\columnwidth]{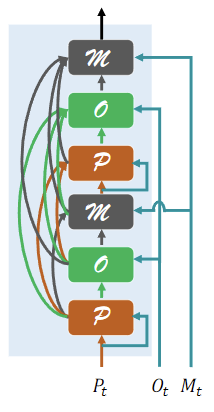}}
        \label{fig:review-7-a}
    }
    \quad
    \subfigure[Parallel IA]{
        \textsf{\includegraphics[width=0.45\columnwidth]{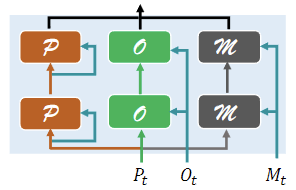}}
        \label{fig:review-7-b}
    }
    \vspace{\BeforeCaptionVSpace}
    \caption{Architecture of the dense serial IA and parallel IA structures. Figure reproduced from Tang \textit{et al.} (2020), \textit{ECCV}.}
    \label{fig:review-7}
\end{figure}

It should be noted that AlphAction defines Instance Level Features as for each instance (independent human subject or object), i.e., the features $P_t$ and $O_t$ of the person and object extracted separately by detection in \autoref{fig:review-5}. And it defines Temporal Memory Features as the features of human subjects in other frames before and after the current frame, i.e., $M_t = [P_{t-L},~\dots~,~P_t,~\dots~,~P_{t+L}]$ (the total number of frames considered at this point is $2L+1$). In addition, the core of AlphAction's IA network lies in how the interaction blocks are designed and how the interaction blocks are integrated (i.e., how to form the IA structure). As shown in \autoref{fig:review-6}, the structure of the Interaction Block is mainly borrowed from the Transformer Block, which has two inputs, namely query (i.e., $P_t$, features of human subjects) and key/value (i.e., $O_t$, features of objects). Further, in \autoref{fig:review-5}, \autoref{fig:review-6}, and \autoref{fig:review-7}, P-Blocks are responsible for modeling the mutual behavior of humans in the same clip, O-Blocks are responsible for detecting the interrelationships between humans and objects, and M-Blocks detect events with strong logical connections along the temporal dimension~\cite{tang2020asynchronous}. And as shown in \autoref{fig:review-6} and \autoref{fig:review-7}, there are three types of IA structures proposed by AlphAction, namely Serial IA, Dense Serial IA, and Parallel IA, which correspond to the different training models used in the later sections.

\subsection{Summary}
\label{ch:review-4}
According to this literature review, the need for smart wards in modern hospitals is increasing and the development of smart wards is beneficial for both healthcare professionals and patients. To the best of the authors' knowledge, there is no method for intelligent recognition of patient actions in the ward. Therefore, in order to meet the need for further upgrading of smart wards and the need for care in areas with increasing aging or relative lack of medical resources, there is a need to develop a method that can identify abnormal patient actions in wards. The development and implementation of this method will help to reduce the human resources needed in each ward area, as well as to reduce the time window for intervention by healthcare professionals after non-medical accidents in the ward for children or senior patients.

In addition, among the current diverse human subject detection methods, more information can be obtained more accurately using human pose estimation methods such as OpenPose. For example, by identifying human skeletal key points and facial key points, we can obtain the smallest rectangular enclosing box more accurately with skeletal key point information, and use facial key points for pre-processing or post-processing of facial blurring to protect the patient's privacy.

Meanwhile, the use of DNN such as AlphAction's IA network and AMU algorithm to form a human action recognizer can help to achieve action recognition with better classification performance. However, as described in \autoref{ch:intro}, we found that the YOLOv3 model, which provides human subject location information to AlphAction's feature extractor, is not accurate enough for some common scenarios in wards. This directly leads to the inability of its efficient and accurate AIA classifier to perform action recognition for such subjects that require attention. Therefore, we use OpenPose as a replacement for YOLOv3 and further train the AIA classifier. This part of the work will be described in detail in the next section.

\section{Methods}
\label{ch:methods}

\begin{figure}[htb]     
\centerline{\includegraphics[width=\columnwidth]{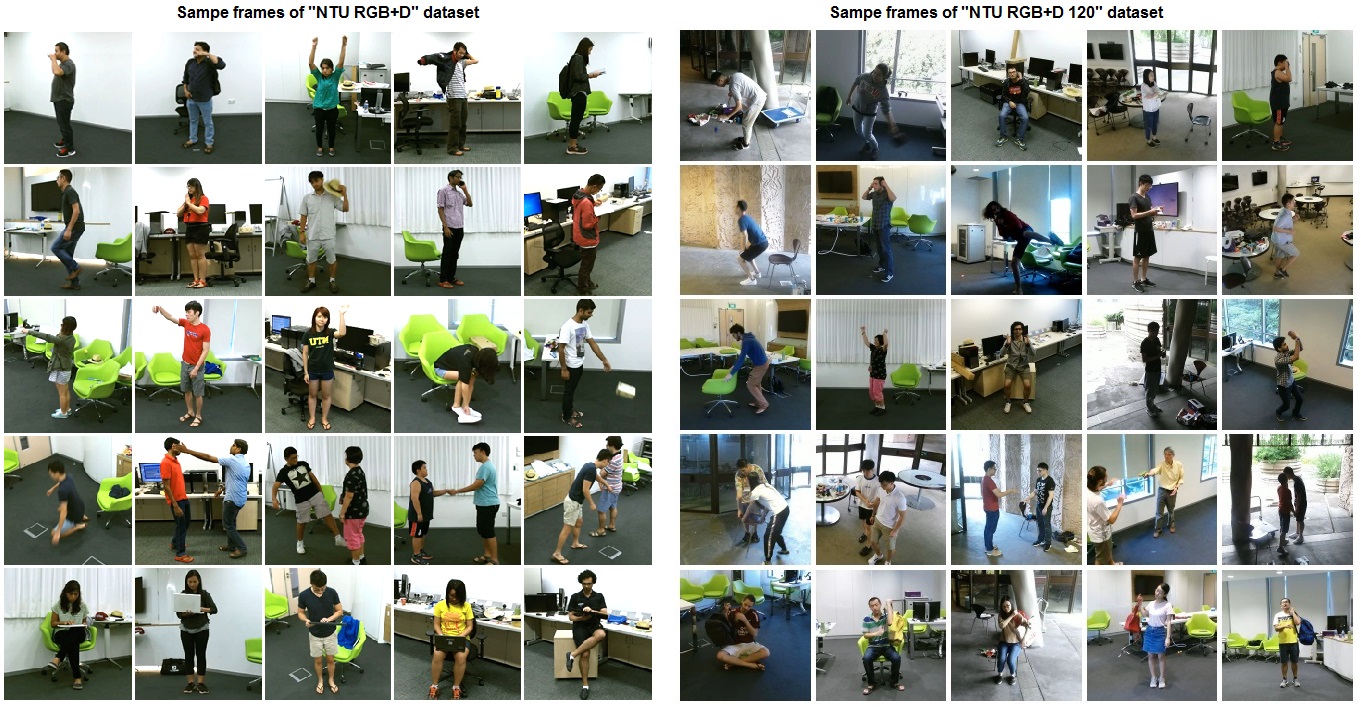}}
\vspace{\BeforeCaptionVSpace}
\caption{Example data provided by the NTU RGB+D/NTU RGB+D 120 datasets. Note that not all the corresponding videos are used in the consequent training process. Figure reproduced from Shahroudy \textit{et al.} (2016), CVPR and Liu \textit{et al.} (2019), TPAMI.} 
\label{fig:methods-0} 
\end{figure}

In this section, we present the implementation of using OpenPose instead of YOLOv3 as a human detector and the general architectural implementation of PoseAction. This is followed by a description of the training and validation set partitioning and preprocessing for the training of the AIA module using NTU RGB+D/NTU RGB+D 120 datasets. Example data from such datasets is shown below in \autoref{fig:methods-0}. Finally, the model evaluation metrics we used are introduced.

\subsection{Architecture and Implementation of PoseAction}
In brief, PoseAction's action recognition module follows the AIA architecture of AlphAction. However, we replaced part of its original feature extractor with an OpenPose-based human subject detector. In this section, we will first describe how to extract the smallest rectangular bounding box using OpenPose, and then how to integrate it into AlphAction's original architecture to form PoseAction.

\subsubsection{From Key Points to Smallest Rectangular Bounding Boxes of Human Subjects}
Similar to what is shown in \autoref{fig:review-2}, OpenPose was also originally designed to perform multi-person 2D pose/key points estimation. In simple terms, these key points will contain the head, torso, and limbs of human subjects. It happens that the human subject's actions are also realized by the motion of these parts, i.e., we only need to specify which parameters AlphAction receives back from the YOLOv3 model, and we can replace it with OpenPose.

After analyzing the publicly available code of AlphAction (\href{https://github.com/MVIG-SJTU/AlphAction}{https://github.com/MVIG-SJTU/AlphAction}), we found that YOLOv3 returns the position information of human subjects in 2D video frames in the form of a set of multidimensional matrices. Its general format is shown in \autoref{eq:1}, which describes exactly the smallest rectangular enclosing box (\textit{bbox}) of each subject and the corresponding subject number. Each row stores the coordinates of the \textit{bbox} and its corresponding subject number. \textit{bbox} coordinates are represented as a quadratic vector, with $x$ and $y$ representing the horizontal and vertical coordinates of its upper-left corner point, while $l$ and $w$ represent the height and width of the \textit{bbox}. The object number increases from $0$ to the last subject $i$.

\begin{equation}
\label{eq:1}
    \begin{bmatrix}
    [[x_0,~y_0,~l_0,~w_0],~0];\\ 
    [[x_1,~y_1,~l_1,~w_1],~1];\\ 
    [[x_2,~y_2,~l_2,~w_2],~2];\\ 
    \dots\\ 
    [[x_i,~y_i,~l_i,~w_i],~i]
    \end{bmatrix}
\end{equation}

\begin{figure}[!t]     
\centerline{\includegraphics[width=0.9\columnwidth]{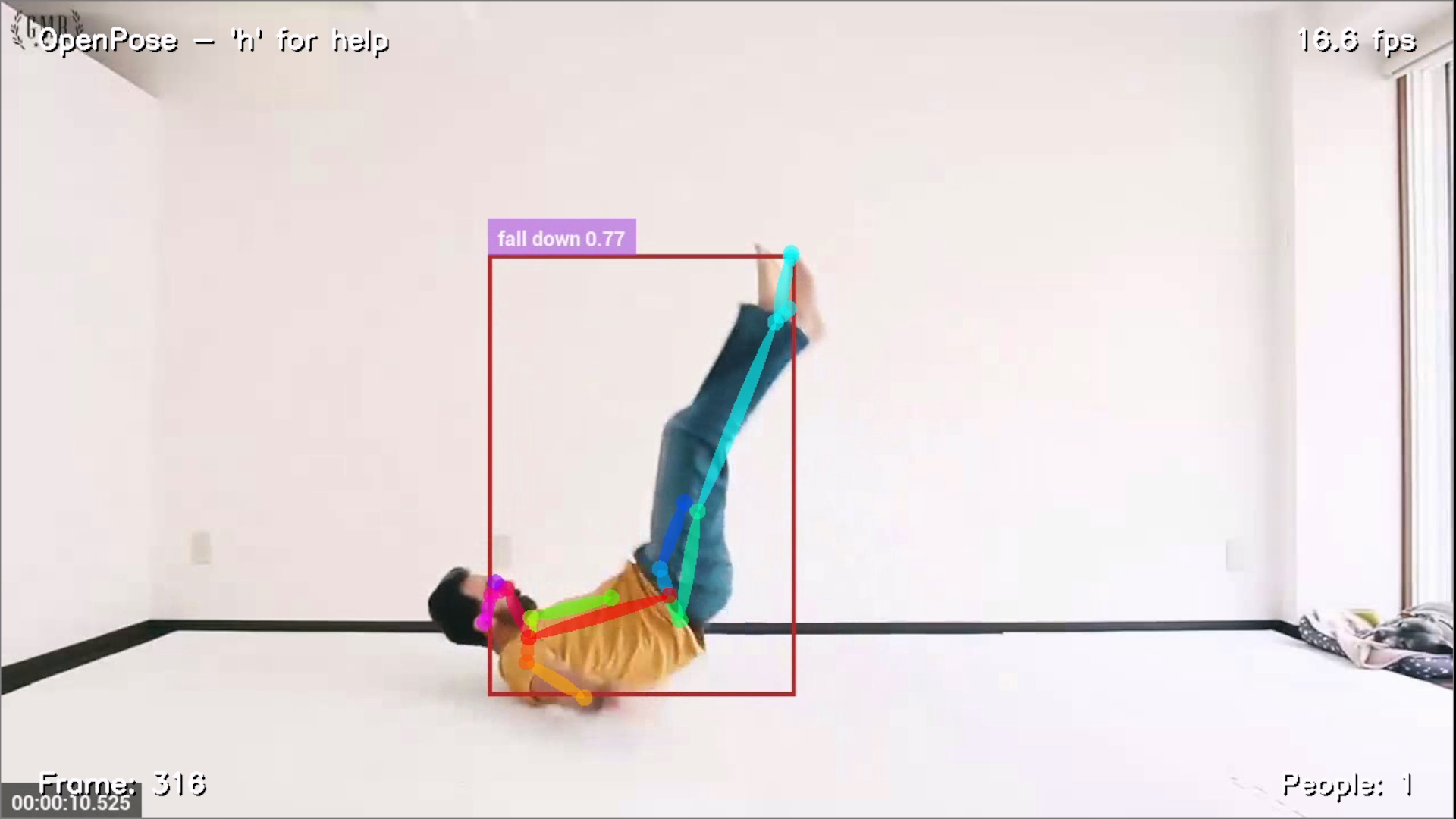}}
\vspace{\BeforeCaptionVSpace}
\caption{This figure shows the human subject key points generated with OpenPose and the smallest rectangular bounding box (\textit{bbox}) obtained using our methods, as well as the results of the action recognition. The original video was derived from \href{https://www.youtube.com/watch?v=ZVzzJ4xDgoE}{YouTube}~\cite{GMB_Fitness_2019}.} 
\label{fig:methods-1} 
\end{figure}

Therefore, the problem we need to solve is how to transform several pairs of $X$ and $Y$ coordinates of key points generated by the OpenPose into the form shown in \autoref{eq:1}. Here, we use the OpenCV2 function \texttt{cv2.rectangle()} for this purpose. This function takes a number of pairs of $X$ and $Y$ coordinates passed in (\autoref{eq:2} is a simplified representation of this stored form) to obtain the \textit{bbox} parameters in $[x,y,l,w]$ format. As a result, we obtain data using OpenPose that can replace the \textit{bbox} of human subjects passed into the AIA module by using YOLOv3 (as shown in \autoref{fig:methods-1}).

\begin{equation}
\label{eq:2}
    [X_0,Y_0;~X_1,Y_1;~X_2,Y_2;~\dots~;~X_j,Y_j]
\end{equation}

It should be noted that here we do not optimize the parameters of OpenPose or perform additional training, because its performance with the default parameters is sufficient for the detection work performed in real-time.

\subsubsection{PoseAction in Detail}
\begin{figure}[!t]     
\centerline{\includegraphics[width=\columnwidth]{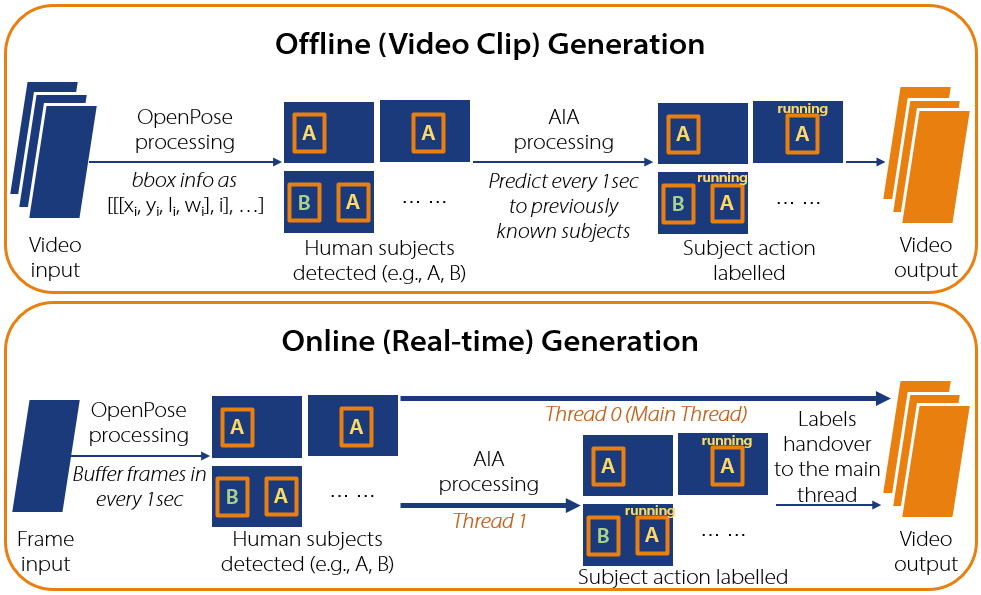}}
\vspace{\BeforeCaptionVSpace}
\caption{Architecture flowchart of the PoseAction.} 
\label{fig:methods-2} 
\end{figure}

For the need for real-time monitoring of patient behavior in the ward, we have developed two modes accordingly, i.e. offline generation mode and online generation mode (\autoref{fig:methods-2}). PoseAction first uses the OpenPose module for human detection and then uses the AIA module for action recognition and fuses the information from both modules for output. The main difference between the offline and online modes is that the offline mode can process each frame of the whole video first (i.e., obtain and buffer all \textit{bbox} in each frame), while the online mode needs to buffer a number of frames in each second for detection and send them to different threads for rendering and further action recognition. 

It should be noted that the input in offline mode is a video clip and its length needs to be greater than 1 second because the AIA module needs to perform action recognition based on the first 1 second of data. In contrast, the input source for online mode can be a local camera or a webcam, which also facilitates its clinical translation in ward action monitoring.

\subsection{Loss and Accuracy During Training}
We recorded the loss and accuracy data based on the training set during the training process at intervals of every 20 iterations and plotted in \autoref{append:trainLossAcc}, where \autoref{append:trainLossAcc-1} shows the loss curves, while \autoref{append:trainLossAcc-2} shows the accuracy curves. It can be seen that both loss and accuracy tend to converge near 100,000 cycles. Therefore, this setting of the maximum training iterations (\texttt{MAX\_ITER}) is reasonable in terms of the balance of resource consumption and results.

\section{Results}
\label{ch:results}

Overall, the training outcome of the AIA module of PoseAction using our preprocessed dataset is very satisfactory. The highest mAP can reach 98.72\% and the lowest mAP can reach 97.90\% with the six different configurations we used for training (\autoref{table:results-1}). Also, by looking at the confusion matrix, it can be seen that the classification accuracy of PoseAction for each class is also quite impressive. In addition, we also tested PoseAction using additional video clips not present in the NTU RGB+D/NTU RGB+D 120 datasets to verify the generalizability of the model and parameters, and the results were also promising.

\begin{table}[!htbp]
\centering
\setlength{\abovecaptionskip}{0pt}    
\setlength{\belowcaptionskip}{10pt}
\caption{PoseAction Performance Comparison on the NTU RGB+D/NTU RGB+D 120 Datasets with Different Configurations}
\setlength{\tabcolsep}{3mm}{
    \begin{tabular}{lllll}
    \toprule[1.5pt]
    \textbf{Configuration} & \textbf{mAP} & \textbf{Precision} & \textbf{Recall} & \textbf{F1 Score} \\
    \specialrule{0em}{1pt}{1pt}
    \hline
    \specialrule{0em}{2pt}{2pt}
    ResNet-50 Baseline     & 0.9840          & 0.9435          & 0.9424          & 0.9426           \\
    ResNet-50 DenseSerial  & 0.9825          & 0.9445          & 0.9421          & 0.9420           \\
    ResNet-50 Parallel     & 0.9790          & 0.9326          & 0.9225          & 0.9224           \\
    ResNet-50 Serial       & 0.9837          & 0.9449          & 0.9443          & 0.9442           \\
    ResNet-101 Baseline    & 0.9825          & \textbf{0.9539} & \textbf{0.9535} & \textbf{0.9536}  \\
    ResNet-101 DenseSerial & \textbf{0.9872} & 0.9532          & 0.9531          & 0.9531           \\
    \bottomrule[1.5pt] 
    \label{table:results-1}
    \end{tabular}
}
\end{table}

\subsection{Experiments on NTU RGB+D/NTU RGB+D 120 Datasets}
As mentioned in previous chapters, the main reason why AlphAction is difficult to directly apply to the caregiving scenario in the ward, besides the lack of performance of its human detector, is that the action labels it contains do not encompass the scene in the ward well. Therefore, we chose the NTU RGB+D/NTU RGB+D 120 datasets to further train the parameters of the AIA model, as they contain more available and critical action classes. Also, due to the short length of the video clips in these datasets, we performed additional preprocessing to meet the minimum requirements for the AIA model training. In this section, we describe the experiment environment for model training and validation, the dataset processing method, and the training and validation setup.

\subsubsection{Experiment Environment}
We used a workstation with Intel Core i9-13900K and Nvidia RTX 3080Ti 12G for training and validation. All data and code were stored and run locally. Also, we used the Ubuntu 22.04.2 LTS operating system to set up the running environment and run all the code. Python version was set to 3.7, and other package dependencies were installed and compiled according to the recommended settings of AlphAction and OpenPose.

\subsubsection{Dataset Formation}
The NTU RGB+D/NTU RGB+D 120 datasets contain 12 desired label classes that were used to generate new train and test subsets~\cite{shahroudy2016ntu,liu2020ntu}. The original NTU RGB+D/NTU RGB+D 120 datasets contain information including RGB videos, depth maps, infrared (IR) data, and 3D skeletons. The labels of the medical-related scenes contained in them and the corresponding number of videos are shown in \autoref{table:methods-1}. Note that we only used RGB videos that contain only 1 subject performing the corresponding action since in a small number of videos there are unrelated (i.e., unlabeled) subjects in the scene.

\begin{table}[!htbp]
    \centering
    \setlength{\abovecaptionskip}{0pt}    
    \setlength{\belowcaptionskip}{10pt}
    \caption{Action Classes, Labels, and Dataset Size for Medical-related Scenarios (RGB Videos) in the NTU RGB+D and NTU RGB+D 120 Datasets}
    \setlength{\tabcolsep}{3mm}{
    \begin{tabular}{llll}
    \toprule[1.5pt]
    \textbf{Action Class} & \textbf{Total} & \textbf{Used} & \textbf{Label} \\
    \specialrule{0em}{1.5pt}{1.5pt} 
    \hline
    \specialrule{0em}{2pt}{2pt}
    \texttt{sneeze/cough}          & 1015                  & 884                  & A041           \\
    \texttt{staggering}            & 1005                  & 893                  & A042           \\
    \texttt{falling down}          & 1040                  & 865                  & A043           \\
    \texttt{headache}              & 1011                  & 887                  & A044           \\
    \texttt{chest pain}            & 1007                  & 893                  & A045           \\
    \texttt{back pain}             & 1009                  & 892                  & A046           \\
    \texttt{neck pain}             & 1012                  & 886                  & A047           \\
    \texttt{nausea/vomiting}       & 1011                  & 886                  & A048           \\
    \texttt{fan self}              & 1007                  & 889                  & A049           \\
    \texttt{yawn}                  & 943                   & 845                  & A103           \\
    \texttt{stretch oneself}       & 997                   & 910                  & A104           \\
    \texttt{blow nose}             & 938                   & 862                  & A105           \\
    \bottomrule[1.5pt]
    \label{table:methods-1}
    \end{tabular}
    }
\end{table}

\begin{figure}[!t]     
\centerline{\includegraphics[width=0.7\columnwidth]{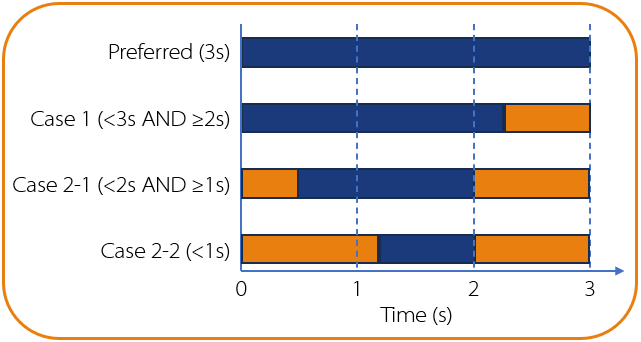}}
\vspace{\BeforeCaptionVSpace}
\caption{An illustration of the preferred length of a video for training and validation and the length of videos in real cases. Blue: the original length of the video; Orange: the extended length of the video.} 
\label{fig:methods-3} 
\end{figure}

Furthermore, since the RGB videos in the original NTU RGB+D/NTU RGB+D 120 datasets varied in length by around 1-3 seconds, we used an additional method to process them. Specifically, as shown in \autoref{fig:methods-3} and \autoref{Methods:algo:1}, we divided and processed all the videos used to build the training and validation sets into two major categories. One category is longer than 2 seconds, and the other category is less than 2 seconds long. For the former, we treat its last frame as a still frame and fill it backward to 3 seconds long. For the latter, we aligned the last frame to the end of the 2$^{\mathrm{nd}}$ second, then the first and the last frames were aligned forward and backward to the 3-second length, respectively.

\begin{algorithm}[!t]
\AlgoFontSize
\DontPrintSemicolon

\KwGlobal{length of the given video clip in frames $\mathcal{T}$}
\KwGlobal{first frame of the given video clip $\mathcal{F}$}
\KwGlobal{last frame of the given video clip $\mathcal{L}$}
\KwGlobal{frames per second of the given video clip $\mathcal{FPS}$}
\BlankLine

\SetKwFunction{fVideoExtension}{VideoExtension}

\KwIn{raw video clip $V_{in}$}
\KwOut{extended video clip $V_{out}$}
\Proc{\fVideoExtension{$V_{in}$}}{
  \uIf{$\mathcal{T} \geq 3$}{
    \Return $V_{in}$\;
  }
  \Else{
  \uIf{$\mathcal{T} \geq 2$}{
    $V_{out}(0:\mathcal{T}-1) \gets V_{in}$\;
    $V_{out}(\mathcal{T}:3\times \mathcal{FPS}-1) \gets \mathcal{L}$\;
  }
  \Else{
    $V_{out}(0:2\times \mathcal{FPS}-\mathcal{T}-1) \gets \mathcal{F}$\;
    $V_{out}(2\times \mathcal{FPS}-\mathcal{T}:2\times \mathcal{FPS}-1) \gets V_{in}$\;
    $V_{out}(2\times \mathcal{FPS}:3\times \mathcal{FPS}-1) \gets \mathcal{L}$\;
  }
  \Return $V_{out}$\;}
}
\caption{Pseudo code of video preprocessing for the training and validation sets.}
\label{Methods:algo:1}
\end{algorithm}

At the same time, the original $1920\times 1080$ 30 FPS video was compressed to $640\times 360$ 25 FPS using the tool (\texttt{process\_ava\_videos.py}) provided by the AlphAction project, and the middle frame was captured as a keyframe for the extraction of subject position information. After that, the video was cut into 3 segments of 1 second each and stored. The extraction of subject position information was performed using the OpenPose module. The information is stored in another form of \textit{bbox}, i.e., the upper left corner coordinates ($x_1,\ y_1$) and the lower right corner coordinates ($x_2,\ y_2$) of the smallest rectangular bounding box. Afterwards, the position information of the subject ($x_1,\ y_1,\ x_2,\ y_2$), the label of actions (i.e., A041-A049 and A103-A105), and the pre-processed video name (e.g., S001C001P001R001A041) are summarized and stored as the dataset annotation for subsequent model training. Finally, these annotations are formatted into COCO-style to meet the input requirements for training using another tool (\texttt{csv2COCO.py}) provided by the AlphAction project.

\subsubsection{Training and Validation}
When dividing the training and validation sets, we split the videos of each action class in a ratio of 8:2, which is determined according to the general requirements in the field of CV.

For the datasets we use and other requirements, we need to further modify the code and the configuration files for training and validation. The modified code and the model parameters we finally obtained will be compiled and uploaded to GitHub later. During the configuration file modification, we mainly modified some common parameters such as the base learning rate (see \autoref{table:methods-2} for details). These parameters affect learning efficiency and are limited by the performance of our workstation. It is important to note that these parameters may not be optimal due to time constraints. In addition, depending on the IA structure and the backbone used, we used a total of six configurations for training, which are: ResNet50-Baseline, ResNet50-DenseSerial, ResNet50-Parallel, ResNet50-Serial, ResNet101-Baseline, and ResNet101-DenseSerial. And it should be noted that the two configurations used as a baseline use the SlowFast network as the backbone, while the rest use the pre-trained 3D CNN backbone pre-trained on the Kinetics-700 dataset provided by the AlphAction authors.

\begin{table}[!htbp]
    \centering
    \setlength{\abovecaptionskip}{0pt}    
    \setlength{\belowcaptionskip}{10pt}
    \caption{Key Parameter Settings for Config Profiles during Training and Validation}
    \setlength{\tabcolsep}{3mm}{
    \begin{tabular}{ll}
    \toprule[1.5pt]
    \textbf{Parameter Name} & \textbf{Value} \\
    \specialrule{0em}{1.5pt}{1.5pt} 
    \hline
    \specialrule{0em}{2pt}{2pt}
    \texttt{NUM\_CLASSES}          & 12 \\
    \texttt{BASE\_LR}              & 0.000125 \\
    \texttt{STEPS}                 & (560000, 720000) \\
    \texttt{MAX\_ITER}             & 100000 \\
    \texttt{VIDEOS\_PER\_BATCH}    & 2 \\
    \bottomrule[1.5pt]
    \label{table:methods-2}
    \end{tabular}
    }
\end{table}

During the training process, we saved the model parameters every 10,000 iterations and performed a validation based on the current model parameters to obtain the according classification accuracy. Finally, we also performed additional validation using the final model parameters obtained after training to obtain a more comprehensive performance evaluation metric. In this work, we used Mean Average Precision (mAP, the most commonly used evaluation metric for object detection problems), Average Precision (AP), Precision, Recall (or Sensitivity), and F1 Score for performance evaluation. Among them, mAP is calculated on the basis of spatial IoU (Intersection over Union) $\geq$ 0.5. It should be noted that AP is the mean value of the highest precision under different recalls, while mAP is the mean value of AP. For a binary classification problem, the precision, recall, and F1 score are calculated as shown in \autoref{eq:3}, \autoref{eq:4}, and \autoref{eq:5}, where $TP$ stands for True Positive, $FP$ stands for False Positive, and $FN$ stands for False Negative.

\begin{equation}
\label{eq:3}
    \mathrm{Precision} = \frac{TP}{TP+FP}
\end{equation}

\begin{equation}
\label{eq:4}
    \mathrm{Recall} = \frac{TP}{TP+FN}
\end{equation}

\begin{equation}
\label{eq:5}
    \mathrm{F1~Score} = \frac{2\times \mathrm{Precision}\times \mathrm{Recall}}{\mathrm{Precision}+\mathrm{Recall}}
\end{equation}

Furthermore, for a multiclass classification problem, we need to calculate the precision and recall for each class before we can get its overall precision and recall. in this work, we use the Macro-average method to calculate the overall precision and recall, because the amount of data for each of our classes is similar and it is possible to evaluate each class equally with this method. The Macro-precision, Macro-recall, and Macro-F1 Score are calculated as shown in \autoref{eq:6}, \autoref{eq:7}, and \autoref{eq:8}, where $P_i$ represents the precision computed on class $i$ and $R_i$ represents the recall computed on class $i$.

\begin{equation}
\label{eq:6}
    \mathrm{Macro\text{-}precision} = \frac{P_1+P_2+\dots+P_i}{i}
\end{equation}

\begin{equation}
\label{eq:7}
    \mathrm{Macro\text{-}recall} = \frac{R_1+R_2+\dots+R_i}{i}
\end{equation}

\begin{equation}
\label{eq:8}
    \mathrm{Macro\text{-}F1~Score} = \frac{2\times \mathrm{Macro\text{-}precision}\times \mathrm{Macro\text{-}recall}}{\mathrm{Macro\text{-}precision}+\mathrm{Macro\text{-}recall}}
\end{equation}

In addition, based on the data recorded during the training process, we plotted loss curves and accuracy curves for the training set every 20 iterations, as well as AP and mAP curves for the validation set every 10,000 iterations. Finally, we also plotted the confusion matrices using the results of the validation conducted with the final model parameters to show the classification results for each class of actions more intuitively.

\subsection{Inference During Training}
Another set of data of interest is the validation inference performed every 10,000 iterations during training, which reflects the classification performance of the model on the validation set at the current iteration. \autoref{append:infAcc} shows the AP and mAP curves for inference during training using different configurations. This shows the AP of each class under different training iterations, as well as the overall mAP. It can be seen that the mAP for the validation set converges at 70,000-80,000 iterations for most configurations. Tthe classification performance for 4 to 5 classes is very good for all configurations, which can be almost the whole time or eventually close to 100\% of the AP.

\subsection{Inference with Final Model Parameters}
Since the APs for each class of the validation set using the final model parameters and the mAPs under each configuration are available from the curves in \autoref{append:infAcc}, no additional presentation of this metric will be made here. However, what we need to additionally show and illustrate here are the confusion matrices obtained after the classification test of the validation set using the final parameters. \autoref{append:infCM} shows these confusion matrices. From these matrices, we can see that the two most likely cases of misclassification for all parameters are the identification of \texttt{yawn} as \texttt{blow nose} and the identification of \texttt{blow nose} as \texttt{yawn}. For these two classification errors, we found that they were more concentrated in the DenseSerial and Parallel configurations using ResNet-50, while the incorrect classification was mitigated in the other four configurations. In addition, the two configurations using ResNet-101 were less prone to these two classification errors. Meanwhile, other classification tasks were significantly less likely to have errors.

\subsection{Additional Evaluations}
\begin{figure}[H]
    \centering
    \subfigure[Action recognition using AlphAction]{
        \textsf{\includegraphics[width=0.9\columnwidth]{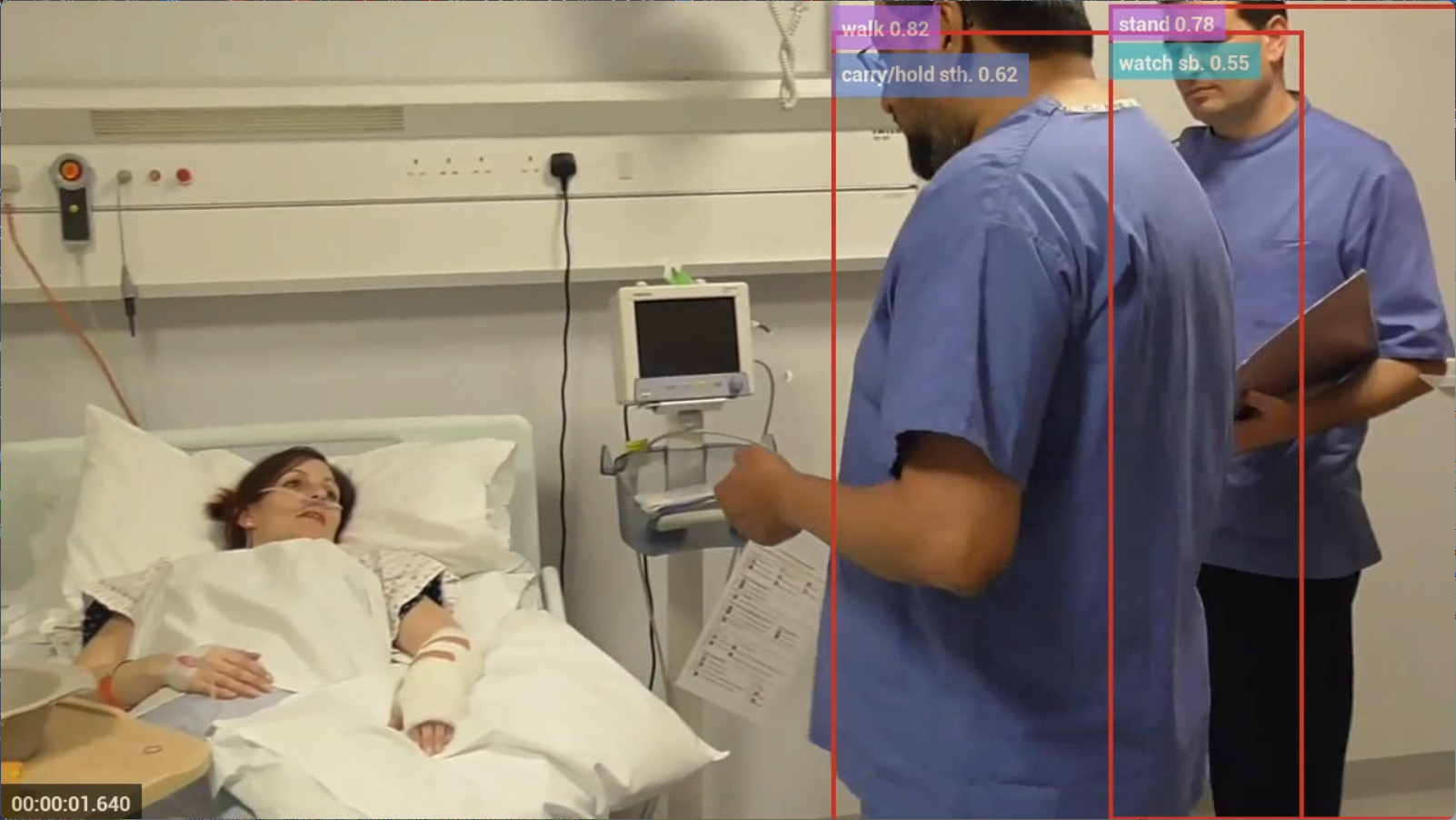}}
        \label{fig:results-27-a}
    }
    \quad
    \subfigure[Action recognition using PoseAction]{
        \textsf{\includegraphics[width=0.9\columnwidth]{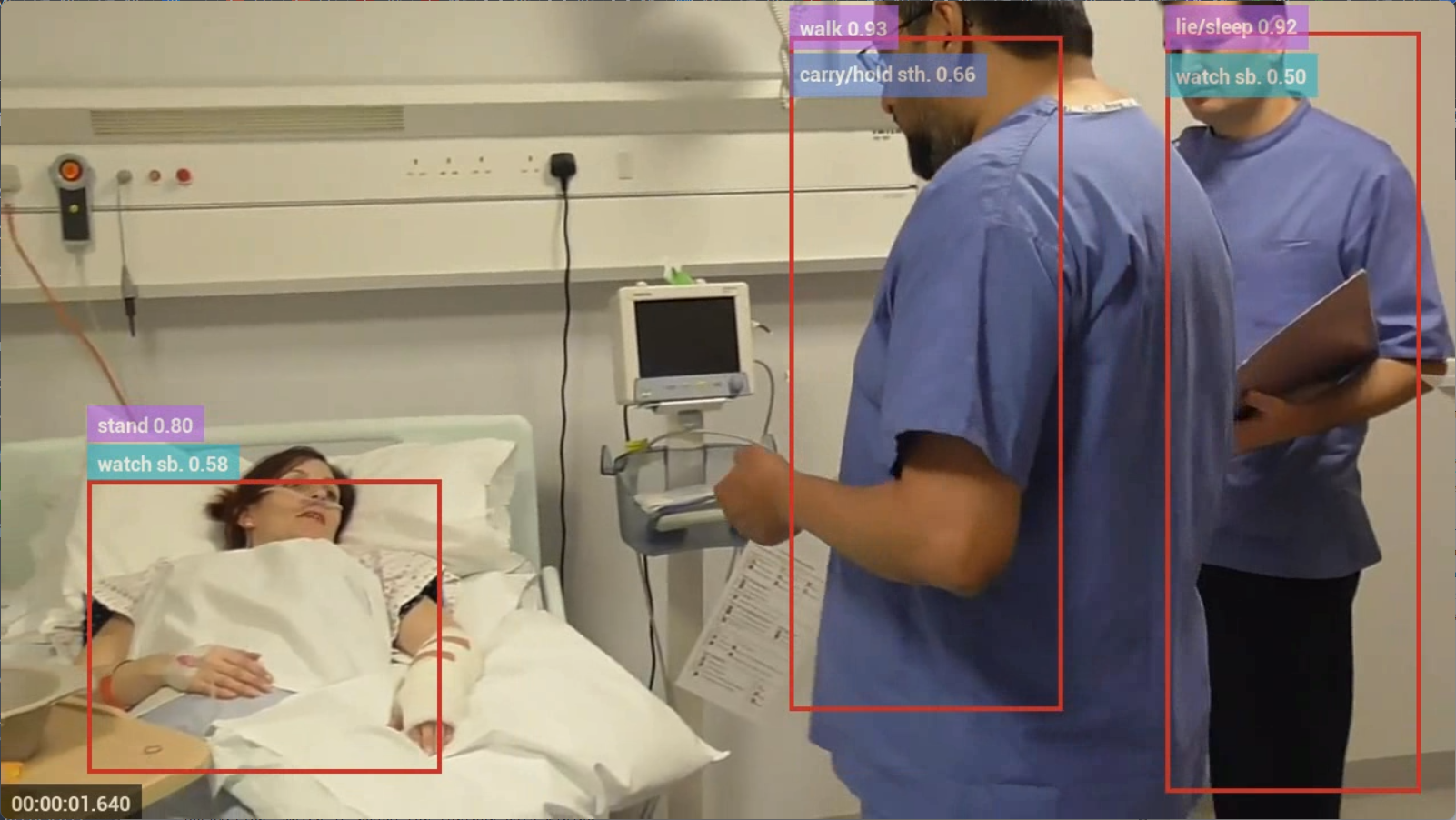}}
        \label{fig:results-27-b}
    }
    \caption{Example 1 - Using AlphAction and PoseAction for action recognition in a real-world ward scene. (a) Only doctors in this scene are detected with AlphAction; (b) The patient is also detected and her actions are well recognized using PoseAction.}
    \label{fig:results-27}
\end{figure}

\begin{figure}[H]
    \centering
    \subfigure[Action recognition using AlphAction]{
        \textsf{\includegraphics[width=0.9\columnwidth]{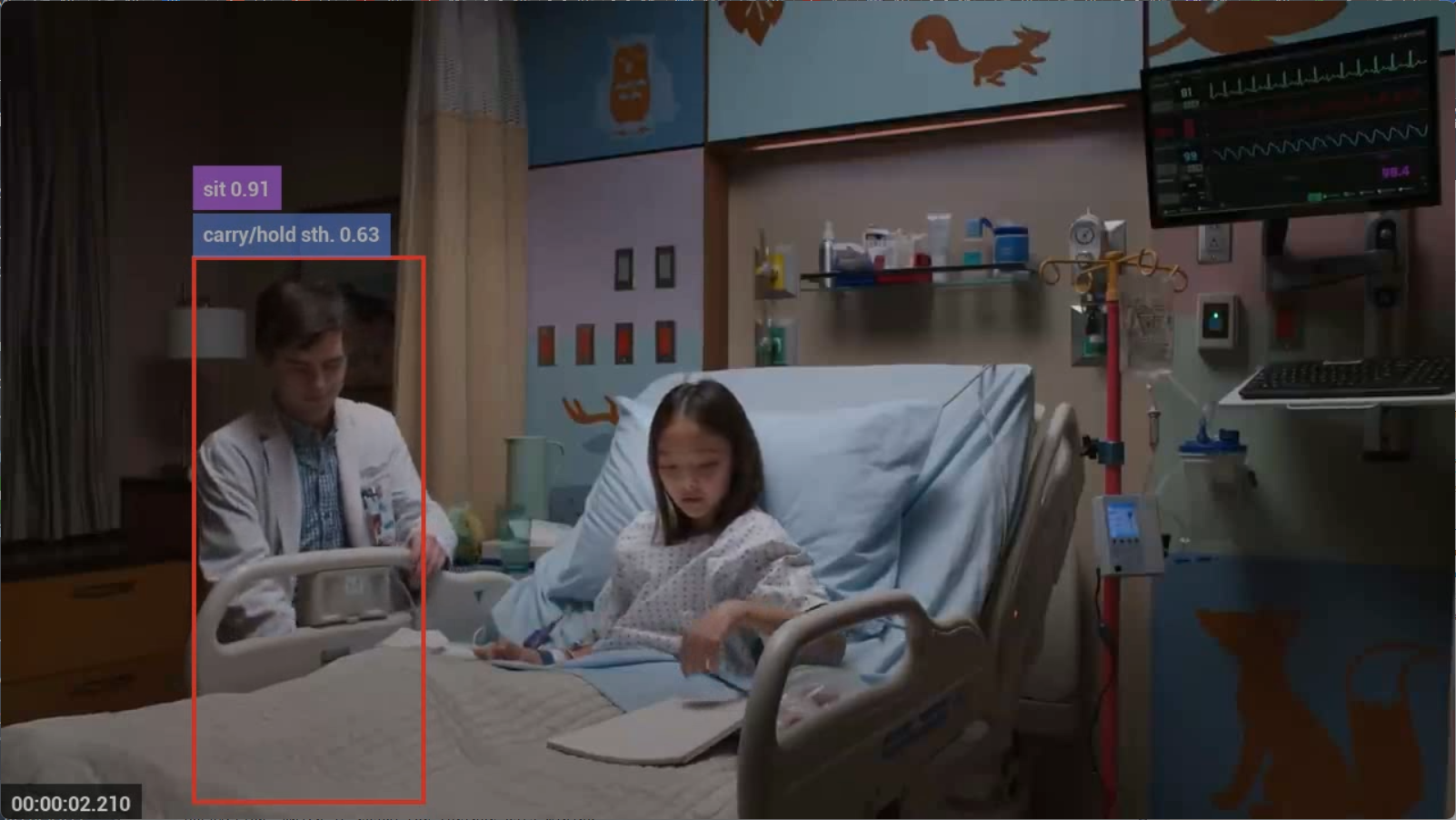}}
        \label{fig:results-28-a}
    }
    \quad
    \subfigure[Action recognition using PoseAction]{
        \textsf{\includegraphics[width=0.9\columnwidth]{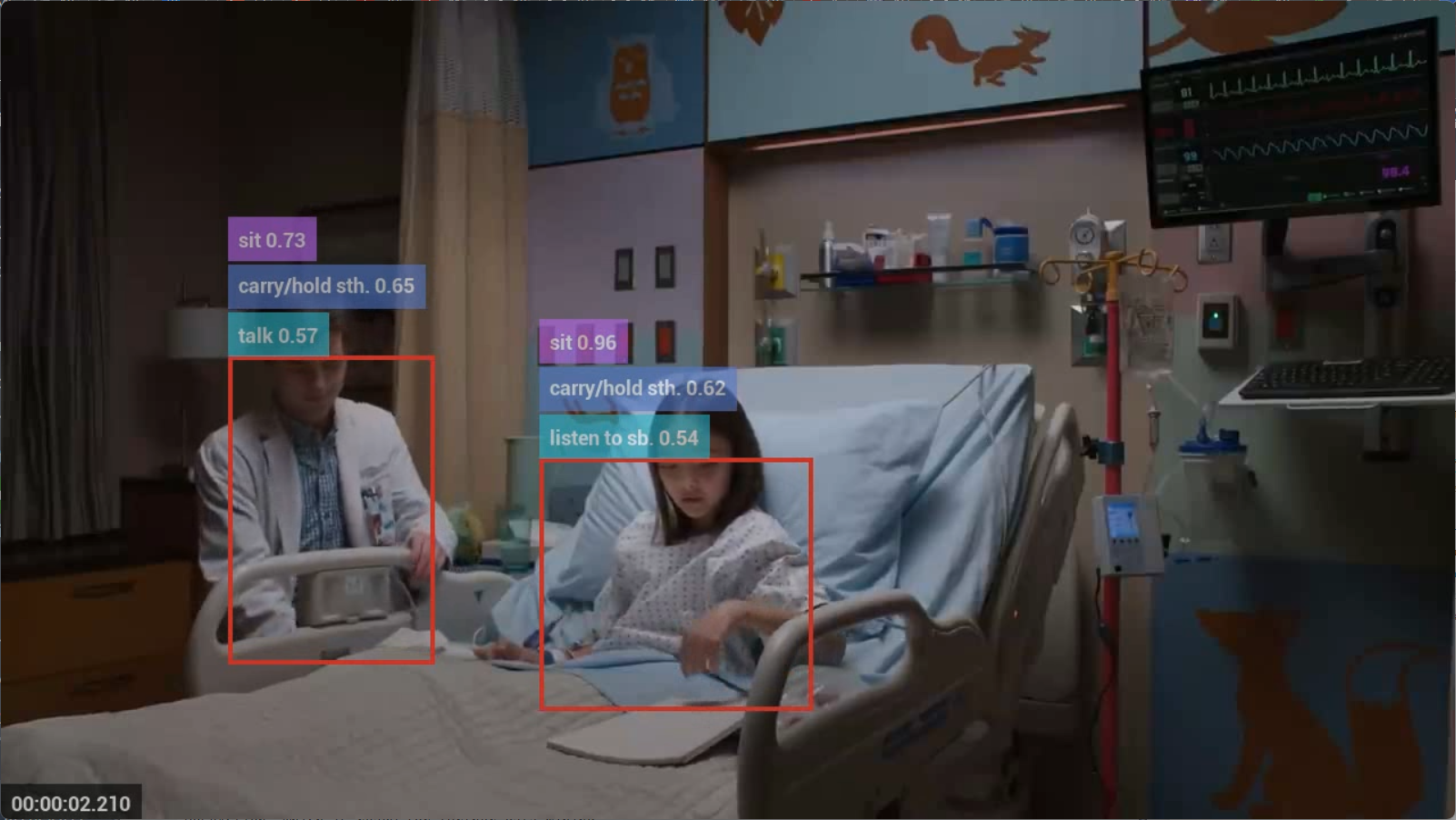}}
        \label{fig:results-28-b}
    }
    \caption{Example 2 - Using AlphAction and PoseAction for action recognition in a TV series ward scene. (a) Only the doctor in this scene is detected with AlphAction; (b) The patient is also detected and her actions are well recognized using PoseAction.}
    \label{fig:results-28}
\end{figure}

Since PoseAction is designed to help hospital wards better care for patients so that medical staff can intervene as soon as possible after an unexpected event, we first tested the AlphAction and PoseAction models using the model weights provided by the AlphAction project. The two short clips tested were both available on YouTube, one was a publicly available clip of a \href{https://www.youtube.com/watch?v=6FAV-s3RYNo}{hospital ward} (\autoref{fig:results-27})~\cite{MEDICINE_in_a_Nutshell_2016}, and the other was a clip from the TV series called \href{https://www.youtube.com/watch?v=lma2NySRJLE}{The Good Doctor} (\autoref{fig:results-28})~\cite{ABC_2022}. As seen in these post-processed video clips, one of the advantages of using PoseAction is that it can detect subjects in complex scenes, such as a patient covered with a quilt. This shows the success of using our OpenPose-based detector, which can be more beneficial for the action recognition of subjects in the ward.

\begin{figure}[H]
    \centering
    \subfigure[Action recognition using AlphAction]{
        \textsf{\includegraphics[width=0.9\columnwidth]{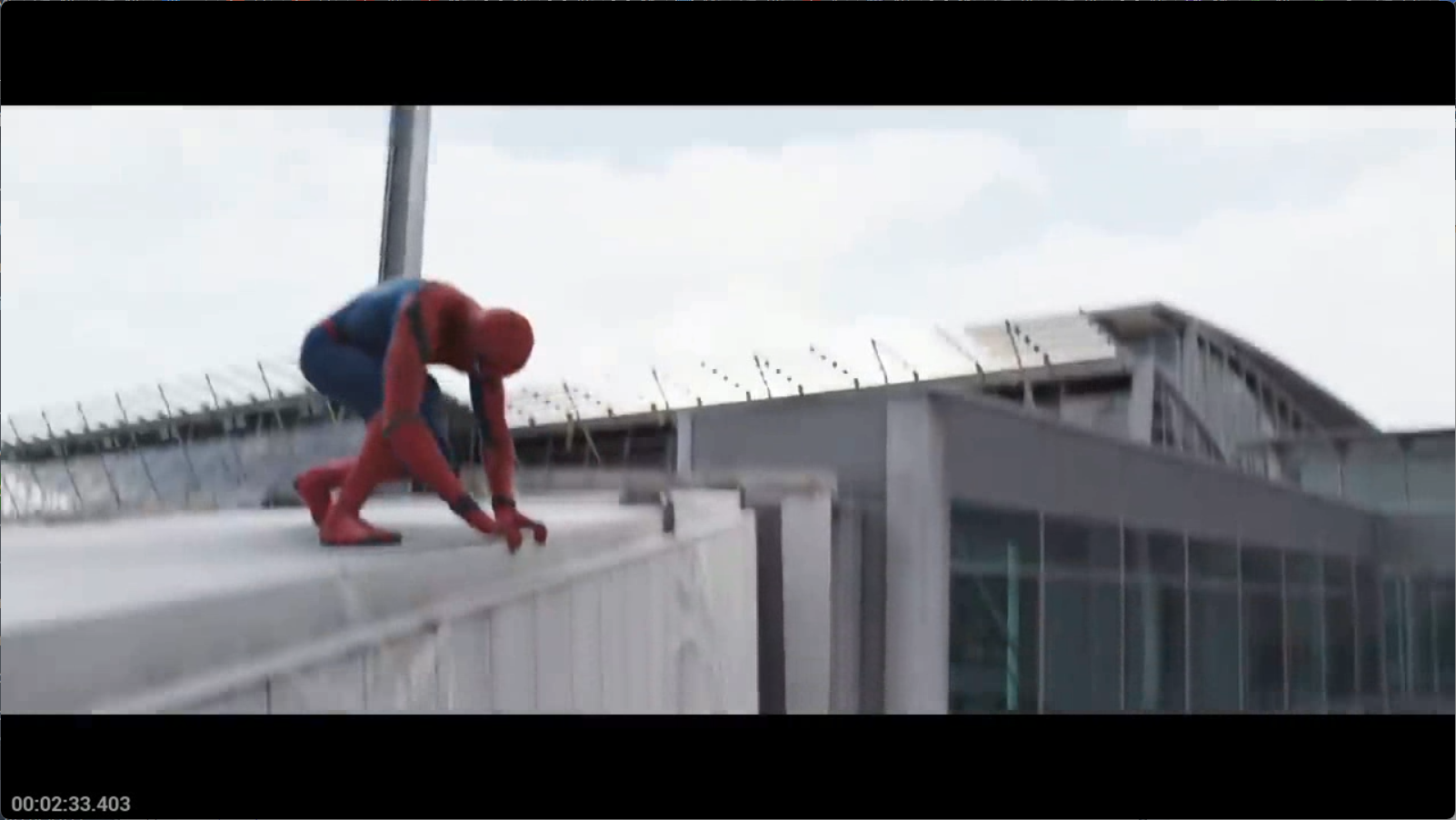}}
        \label{fig:results-25-a}
    }
    \quad
    \subfigure[Action recognition using PoseAction]{
        \textsf{\includegraphics[width=0.9\columnwidth]{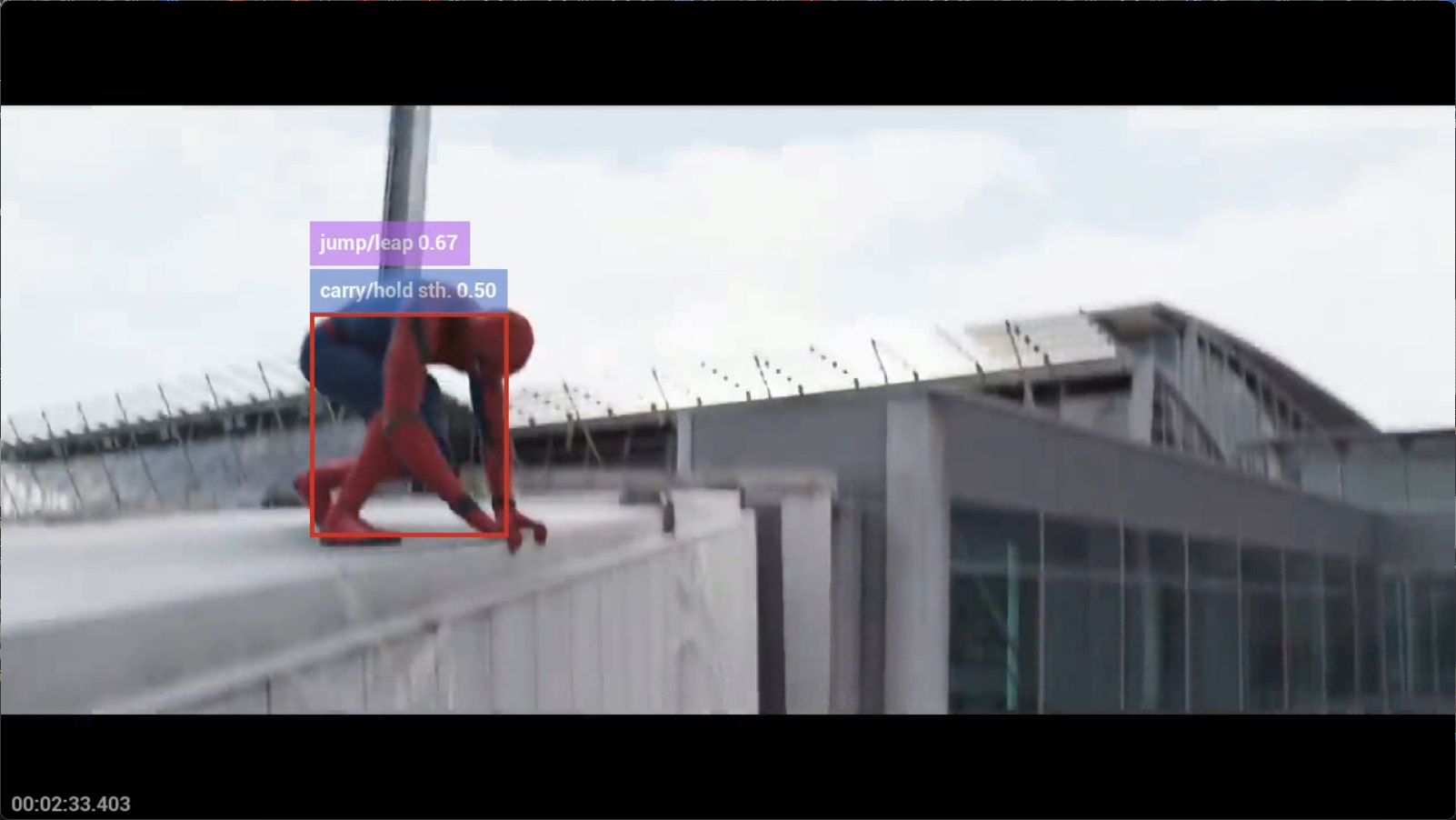}}
        \label{fig:results-25-b}
    }
    \caption{Example 3 - A keyframe comparison of human subject detection and action recognition for a long movie clip.}
    \label{fig:results-25}
\end{figure}

\begin{figure}[H]
    \centering
    \subfigure[Number of subjects detected and actions recognized using AlphAction]{
        \textsf{\includegraphics[width=0.7\columnwidth]{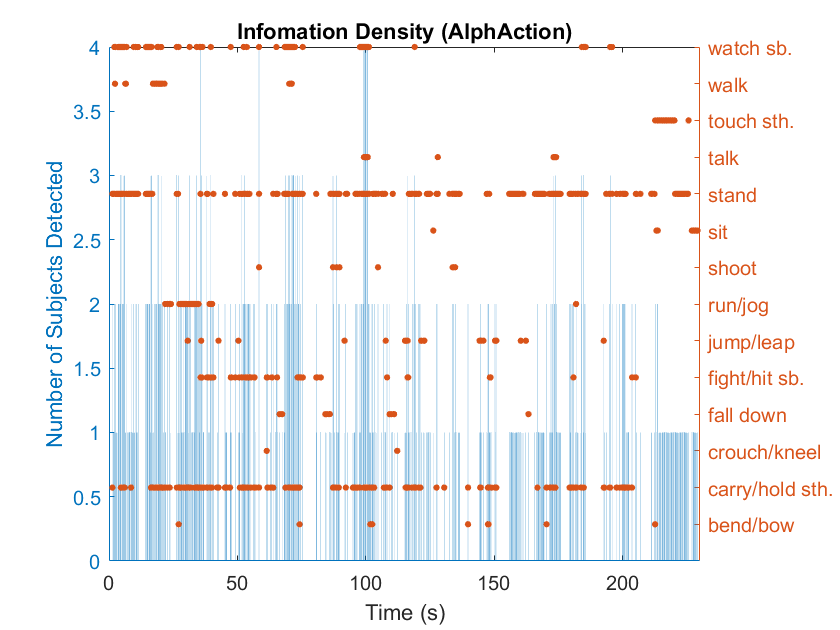}}
        \label{fig:results-26-a}
    }
    \quad
    \subfigure[Number of subjects detected and actions recognized using PoseAction]{
        \textsf{\includegraphics[width=0.7\columnwidth]{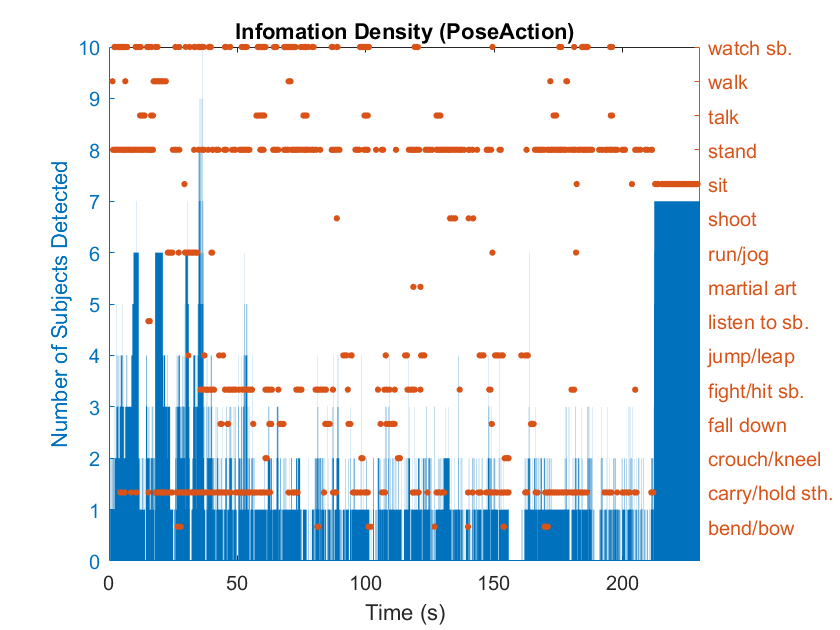}}
        \label{fig:results-26-b}
    }
    \caption{A comparison of information density extraction performance between AlphAction and PoseAction. It shows that more subjects can be detected using the PoseAction model, and therefore more information about subjects' actions can be obtained. Thus, it can be said that PoseAction can better extract the action information of human subjects in the test movie clip.}
    \label{fig:results-26}
\end{figure}

Also, we tested the AlphAction model as well as the proposed PoseAction model using a movie clip (available on \href{https://www.youtube.com/watch?v=cfVY9wLKltA&t=29s}{YouTube}, around 4 minutes in length) containing multiple subjects and complex action scenes with model weights given by the AlphAction project~\cite{TopMovieClips_2019}. A comparison of one set of scenes is shown in \autoref{fig:results-25}, and see \autoref{fig:results-26} for the information density comparison of the whole video. It is obvious that when using AlphAction, the number of subjects that can be successfully detected for most scenes is significantly reduced. Therefore, the amount of action data that can be obtained is also relatively small and it can be said that more information can be extracted from the video using PoseAction, which is ready for further data mining.

\begin{figure}[H]
    \centering
    \subfigure[Case 1]{
        \textsf{\includegraphics[width=0.46\columnwidth]{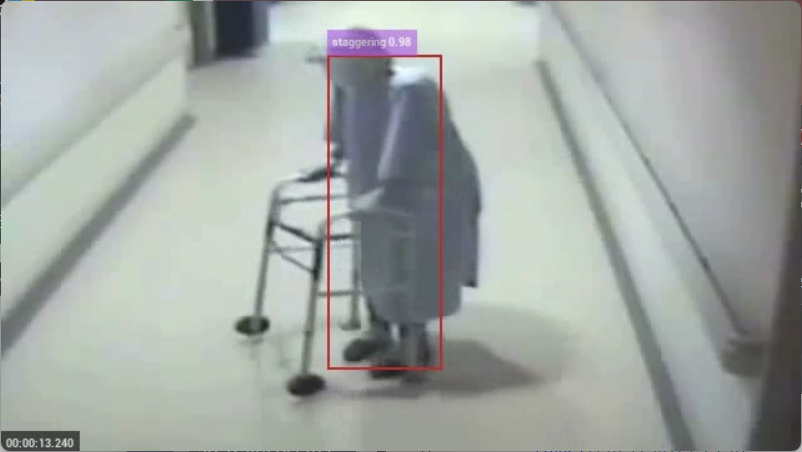}}
        \label{fig:results-29-a}
    }
    \quad
    \subfigure[Case 2]{
        \textsf{\includegraphics[width=0.46\columnwidth]{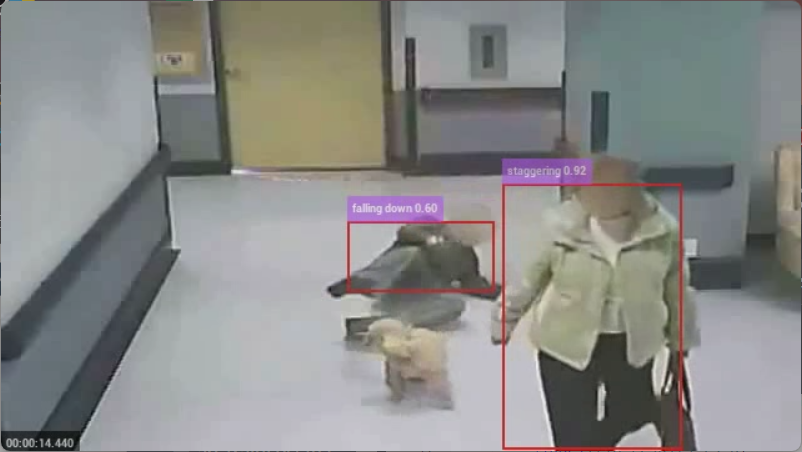}}
        \label{fig:results-29-b}
    }
    \quad
    \subfigure[Case 3]{
        \textsf{\includegraphics[width=0.46\columnwidth]{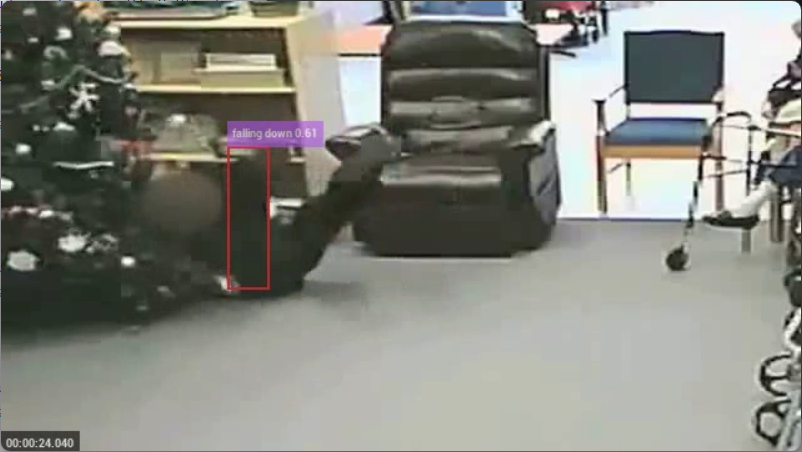}}
        \label{fig:results-29-c}
    }
    \quad
    \subfigure[Case 4]{
        \textsf{\includegraphics[width=0.46\columnwidth]{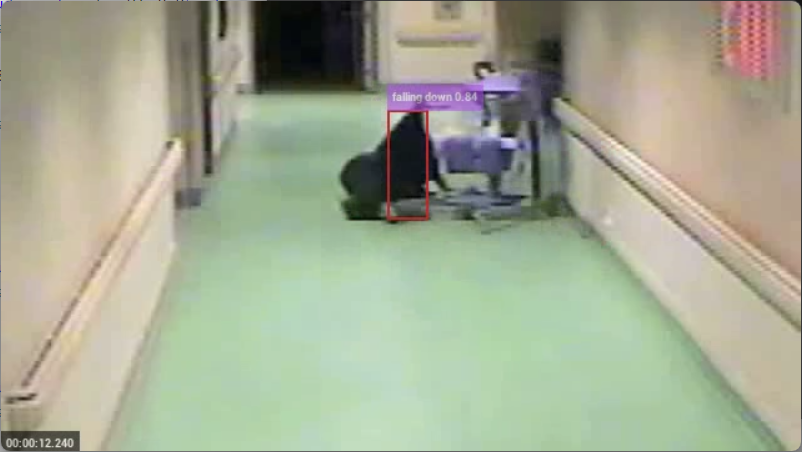}}
        \label{fig:results-29-d}
    }
    \quad
    \subfigure[Case 5]{
        \textsf{\includegraphics[width=0.46\columnwidth]{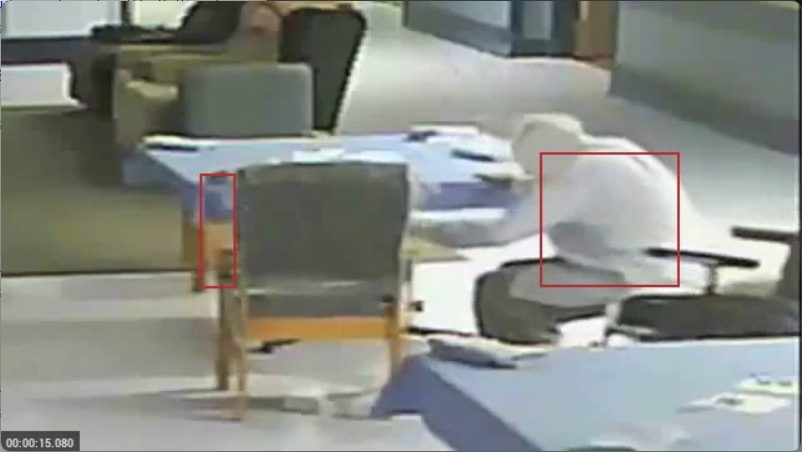}}
        \label{fig:results-29-e}
    }
    \caption{Example 4 - Using PoseAction with our model weights for action recognition in real-world ward scenes. (a) Incorrect weight shift while standing and turning resulting in a fall, \texttt{staggering} recognized; (b) Incorrect weight shift while walking forward resulting in a fall, \texttt{staggering} and \texttt{falling down} recognized; (c) Tripping while walking and turning, \texttt{falling down} recognized; (d) Tripping while walking forward, \texttt{falling down} recognized; (e) Loss of support from an external object while sitting down resulting in a fall, \texttt{falling down} not recognized.}
    \label{fig:results-29}
\end{figure}

In addition, we used video material provided in a study of the circumstances of falls in elderly people residing in long-term care by Robinovitch \textit{et al.} for a more realistic test~\cite{robinovitch2013video}. These videos provide five common scenarios of elderly people falling, such as losing support from external objects while sitting, tripping while walking forward, and tripping while walking forward. The resolution of these videos is 640$\times$480 with frame rates of 4 to 15 FPS, which undoubtedly places high demands on the generalizability of the model and parameters (since the quality of the training data is much higher than these test videos). It should be noted that here we used model weights obtained from training based on the NTU RGB+D/NTU RGB+D 120 datasets. Examples are shown in \autoref{fig:results-29}. From these examples, it can be seen that the two actions of \texttt{staggering} and \texttt{falling down} of the subjects can be well recognized in most cases. Except for the fifth video (Figure \autoref{fig:results-29-e}) in which the subject is not detected continuously because the color of his/her clothes is too close to the background and thus the fall is not recognized successfully, there is no issue that the subjects are not detected and the according actions are not recognized in all other videos.

\section{Discussion}
\label{ch:discussion}

In this study, we introduce and implement the PoseAction model, designed for the purpose of detecting human subjects and recognizing their actions. This model's key advantage lies in its utilization of a novel dataset to identify behaviors prevalent in ward scenarios with potential patient risk. Concurrently, it exhibits superior subject detection performance compared to several existing action recognition models, including AlphAction. In thissection, we will delve deeper into PoseAction's performance, explore potential applications in various domains, and address its current limitations. In conclusion, we express confidence that the PoseAction model will see broader adoption in the future. Given that subject occlusion is not confined to ward scenes, recognizing occluded subjects and actions across a broader array of scenarios poses a significant challenge. Examples encompass action recognition and prediction in densely populated environments. In this context, we present a straightforward yet viable approach to address this issue by employing an integrated model with enhanced performance in each constituent component.

\subsection{Performance Analysis}
In this work, we successfully constructed the PoseAction model that fulfilled our expectations of its functionality and performance. For the training and validation based on the NTU RGB+D/NTU RGB+D 120 datasets, the theoretical performance is undoubtedly significantly improved compared to the original AlphAction. This significant improvement may be attributed to the use of transfer learning for training, i.e., training on the basis of the model weights provided by the AlphAction project allows for faster convergence of loss and accuracy, and better action classification performance. On the other hand, it may be due to the fact that the NTU RGB+D/NTU RGB+D 120 datasets contain much higher quality video clips than the AVA dataset, which allows the model to learn the features of the target action classes better even training after video resolution and frame rate compression (1080P 30FPS $\rightarrow$ 360P 25FPS).

Moreover, the tests on the additional video clips (\autoref{fig:results-27}, \autoref{fig:results-28}, \autoref{fig:results-25}, and \autoref{fig:results-29}) show that both the model weights provided in the AlphAction project and the model weights we trained can be used to obtain better performance on human detection compared to AlphAction. This is certainly critical, since the algorithm cannot recognize the subject's action from an undefined range. In particular, for subjects covered with a quilt in hospital wards (\autoref{fig:results-27} and \autoref{fig:results-28}), the use of our OpenPose-based detector allows for better human detection, thus allowing the AIA module to perform well.

The code for PoseAction, the tools we used to preprocess the NTU RGB+D/NTU RGB+D 120 datasets, the model weights we trained, and the video clips generated during testing will all be uploaded to our GitHub repository once they are organized.

\subsection{Other Possible Applications}
\begin{figure}[H]     
\centerline{\includegraphics[width=0.9\columnwidth]{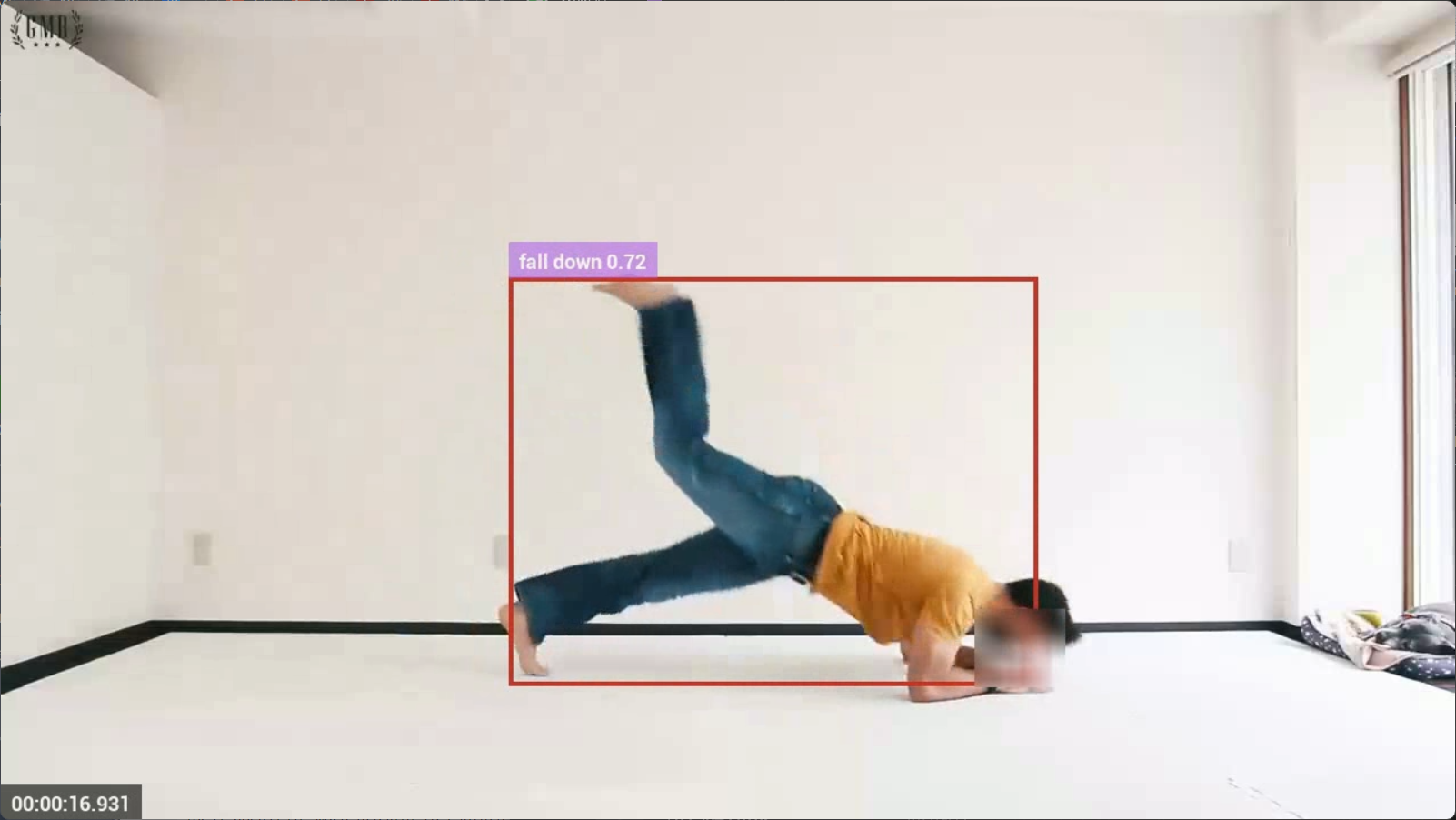}} 
\caption{PoseAction enables real-time face blurring with OpenPose's ability to detect key points on the face of human subjects to protect their privacy. The original video was derived from \href{https://www.youtube.com/watch?v=ZVzzJ4xDgoE}{YouTube}~\cite{GMB_Fitness_2019}.} 
\label{fig:discussion-1} 
\end{figure}

The subject detection module of the proposed PoseAction model, which is derived from the OpenPose model, can also generate data including the skeleton keypoints of subjects as well as the facial keypoints. Such information can be further used for face blurring to ensure the privacy of the subject (\autoref{fig:discussion-1}), or for motion reconstruction and pattern analysis using skeletal keypoint data for patients in wards.

In addition, PoseAction can be considered to assist in determining the mental or physical condition of a patient by tracking behavioral data (e.g., the number and duration of different actions as well as the subject's movement trajectory) over time for the same subject. For example, a subject who has unfortunately been in a car accident a few days ago may show behaviors like a decrease in daily bed rest, or an increase in exercise time, indicating that he/she may be recovering physically. Or, for example, a subject who has had an unfortunate post-operative depression but whose behavior has a decrease in daily sighing, or has an increase in time spent interacting with others, may indicate a recovery in his/her psychological condition. This is one of the next areas of research in which we can use PoseAction to collaborate with hospitals and patients.

\subsection{Limitations}
Although the PoseAction model proposed in this work has excellent performance in many aspects, we also found that it still has some limitations. For example, when using the model weights provided by the AlphAction project, there are still times when some actions are misclassified with high confidence (e.g., the patient in Figure \autoref{fig:results-27-b} is recognized as \texttt{stand} with 0.80 confidence even though she is lying on the bed). Also, when using the weights obtained from our training based on the NTU RGB+D/NTU RGB+D 120 datasets, some common actions (e.g., standing or walking) could not be classified well because we only included labels for 12 common medical-related actions. In addition, as shown in Figure \autoref{fig:results-29-e}, our human detector is also unable to ensure consistent and accurate detection for some heavily occluded subjects, or when the color of the subject's clothing is very close to the background color, resulting in the inability of the AIA module to recognize the subject's actions. This is because the recognition process relies on the action information of the subject during the previous second of the current time point, and thus the action information/recognition is also discontinuous when the same subject cannot be detected continuously.

Besides, due to the public availability restriction of the NTU RGB+D/NTU RGB+D 120 datasets, i.e., the \href{https://rose1.ntu.edu.sg/dataset/actionRecognition/}{ROSE Lab} that built this dataset does not allow us to use this dataset for derivation or generation of new datasets, we have no way to make our train/val dataset public currently. As a result, we hope to further seek collaboration with local hospitals more broadly to capture more action data that actually occurs in the ward area and thus build a public video dataset that serves only action detection in medical-related scenarios.

\section{Conclusion and Future Work}
\label{ch:concl}

In this study, we introduce the PoseAction model, which amalgamates the strengths of AlphAction and OpenPose. This model excels in both human subject detection and action recognition. We fine-tuned the AIA module through transfer learning, employing video clips featuring medical-related scenarios from the NTU RGB+D/NTU RGB+D 120 datasets. The proposed model achieved an impressive 98.72\% mean Average Precision (mAP) at IoU=0.5 on this dataset for action recognition. Additionally, we evaluated PoseAction using publicly available video clips. The test results demonstrate PoseAction's robust generalization, achieving human detection and action recognition in complex scenes. Furthermore, PoseAction outperforms AlphAction in extracting action information from videos. These advantages enable more effective exploitation of human behavior information within videos for subsequent data mining. For instance, tracking specific subjects and analyzing their behavior patterns over time to infer their mental and physical states. This will undeniably benefit intelligent ward care, a necessity in the realm of smart healthcare. Another innovative aspect of this work is the pioneering application of Computer Vision (CV) and Deep Learning (DL) approaches to the field of ward care The potential for medical translation of this project is also a promising avenue for future research, as it can assist medical personnel in early detection of abnormalities in patients' actions and provide timely assistance.

Nevertheless, the training data for PoseAction is currently limited, particularly in terms of label diversity. So, the subsequent step involves utilizing a more diverse dataset (including general actions) to train the model's parameters for improved generalization. In addition, we have also considered manually partially masking the subjects in the training set to further validate the model performance. However, since it is time-consuming and labor-intensive to do so for the large number of videos we used, the author would like to use the skeleton key points generated by OpenPose to perform automated masking in the subsequent work to further verify the performance of PoseAction.

\newpage
\bibliographystyle{unsrtnat}

\newpage
\begin{appendices}
\section{Loss and Accuracy Curves During Training}
\label{append:trainLossAcc}

\subsection{Loss Curves During Training}
\label{append:trainLossAcc-1}
\begin{figure}[H]     
\centerline{\includegraphics[width=0.8\columnwidth]{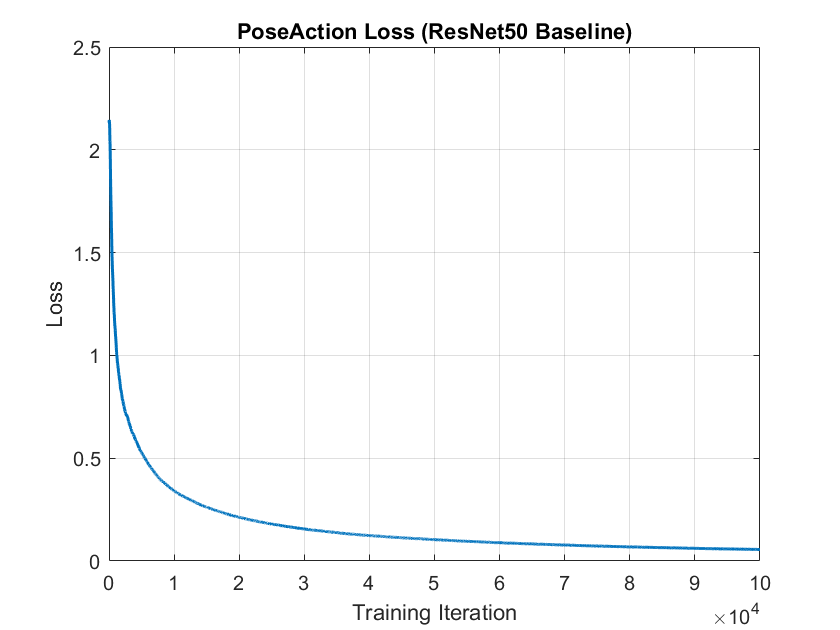}}
\vspace{\BeforeCaptionVSpace}
\caption{PoseAction training loss using the ResNet-50 Baseline configuration.} 
\label{fig:results-1} 
\end{figure}

\begin{figure}[H]     
\centerline{\includegraphics[width=0.8\columnwidth]{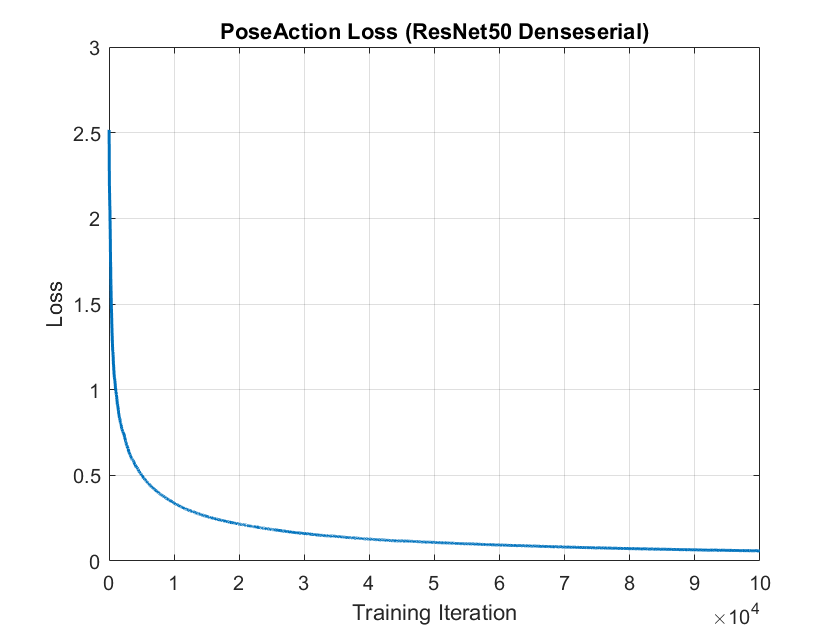}}
\vspace{\BeforeCaptionVSpace}
\caption{PoseAction training loss using the ResNet-50 DenseSerial configuration.} 
\label{fig:results-2} 
\end{figure}

\begin{figure}[H]     
\centerline{\includegraphics[width=0.8\columnwidth]{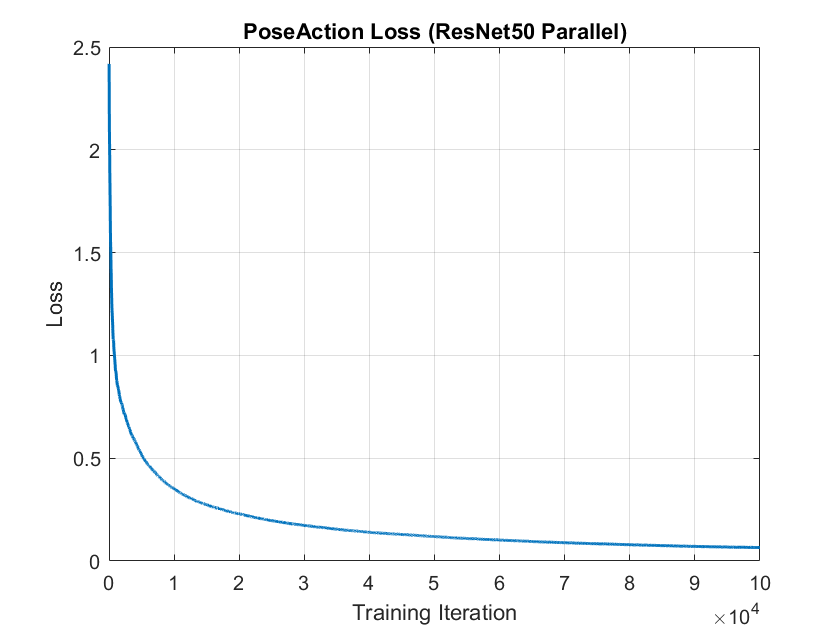}}
\vspace{\BeforeCaptionVSpace}
\caption{PoseAction training loss using the ResNet-50 Parallel configuration.} 
\label{fig:results-3} 
\end{figure}

\begin{figure}[H]     
\centerline{\includegraphics[width=0.8\columnwidth]{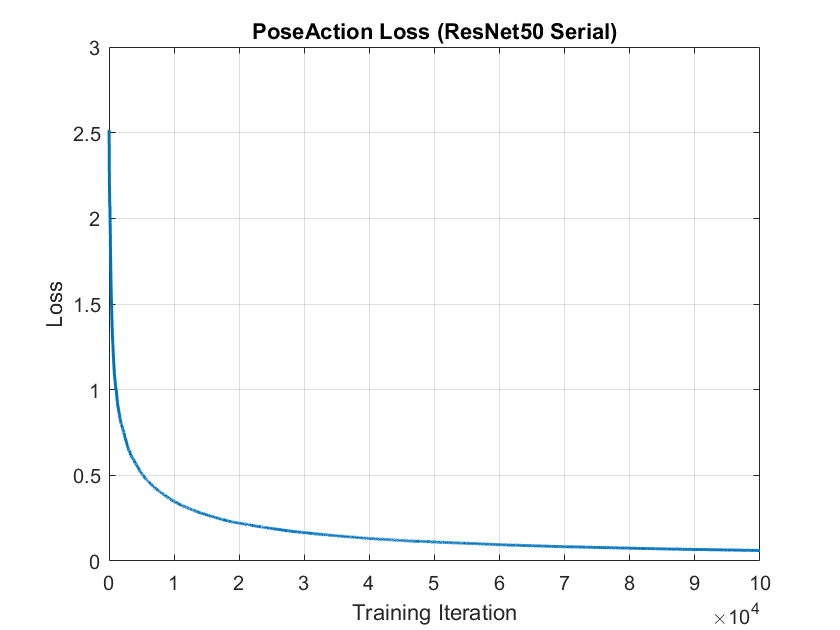}}
\vspace{\BeforeCaptionVSpace}
\caption{PoseAction training loss using the ResNet-50 Serial configuration.} 
\label{fig:results-4} 
\end{figure}

\begin{figure}[H]     
\centerline{\includegraphics[width=0.8\columnwidth]{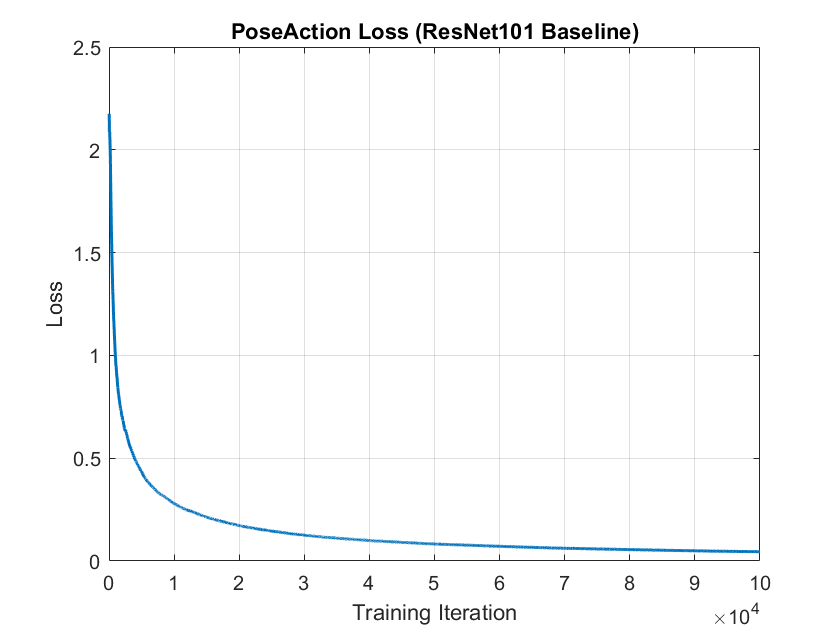}}
\vspace{\BeforeCaptionVSpace}
\caption{PoseAction training loss using the ResNet-101 Baseline configuration.} 
\label{fig:results-5} 
\end{figure}

\begin{figure}[H]     
\centerline{\includegraphics[width=0.8\columnwidth]{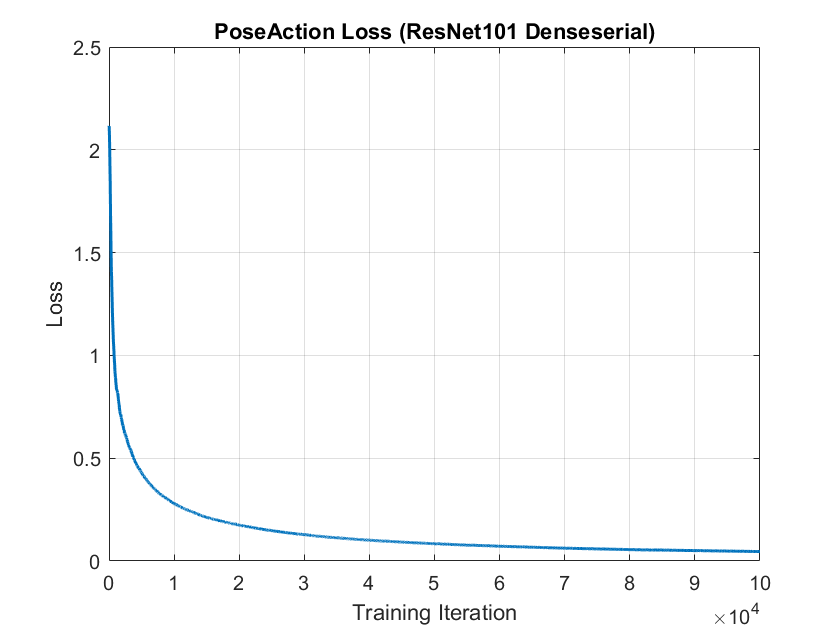}}
\vspace{\BeforeCaptionVSpace}
\caption{PoseAction training loss using the ResNet-101 DenseSerial configuration.} 
\label{fig:results-6} 
\end{figure}

\subsection{Accuracy Curves During Training}
\label{append:trainLossAcc-2}

\begin{figure}[H]     
\centerline{\includegraphics[width=0.8\columnwidth]{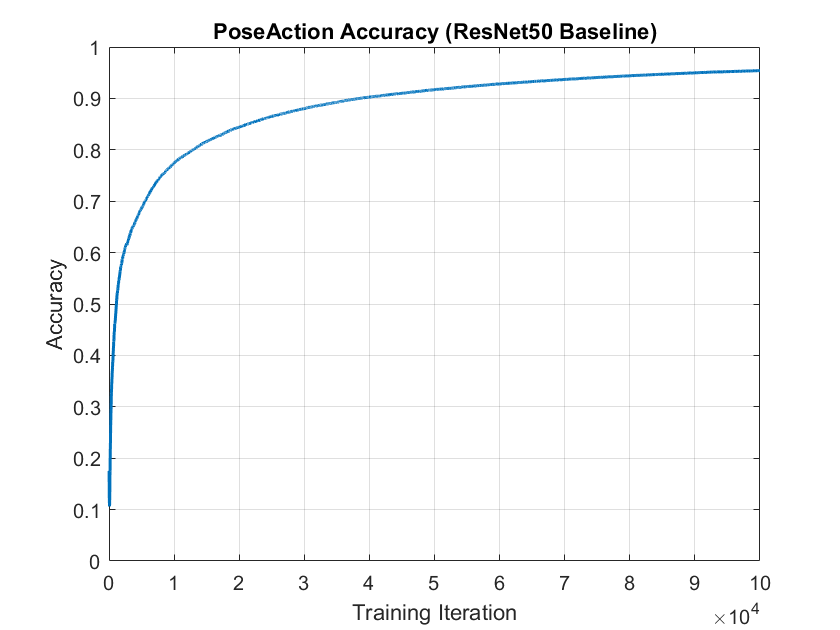}}
\vspace{\BeforeCaptionVSpace}
\caption{PoseAction training accuracy using the ResNet-50 Baseline configuration.} 
\label{fig:results-7} 
\end{figure}

\begin{figure}[H]     
\centerline{\includegraphics[width=0.8\columnwidth]{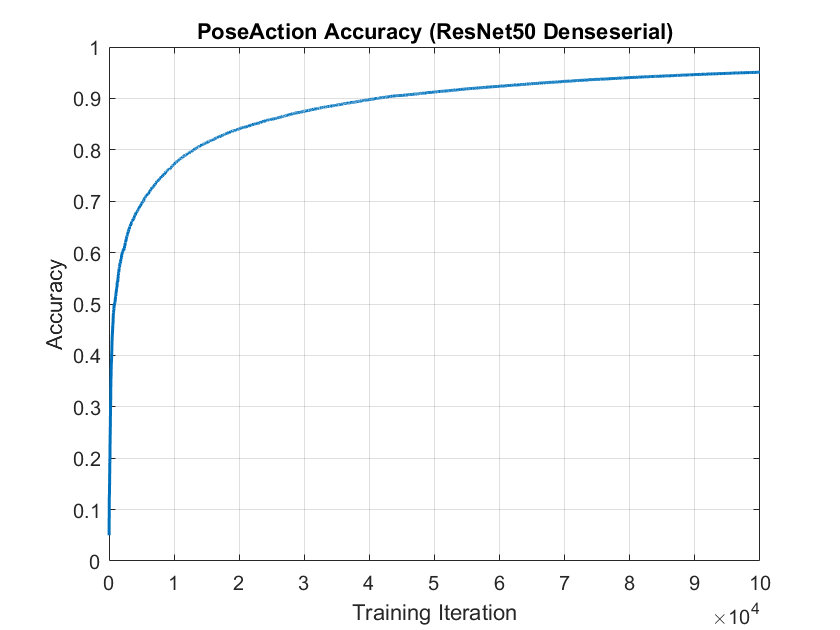}}
\vspace{\BeforeCaptionVSpace}
\caption{PoseAction training accuracy using the ResNet-50 DenseSerial configuration.} 
\label{fig:results-8} 
\end{figure}

\begin{figure}[H]     
\centerline{\includegraphics[width=0.8\columnwidth]{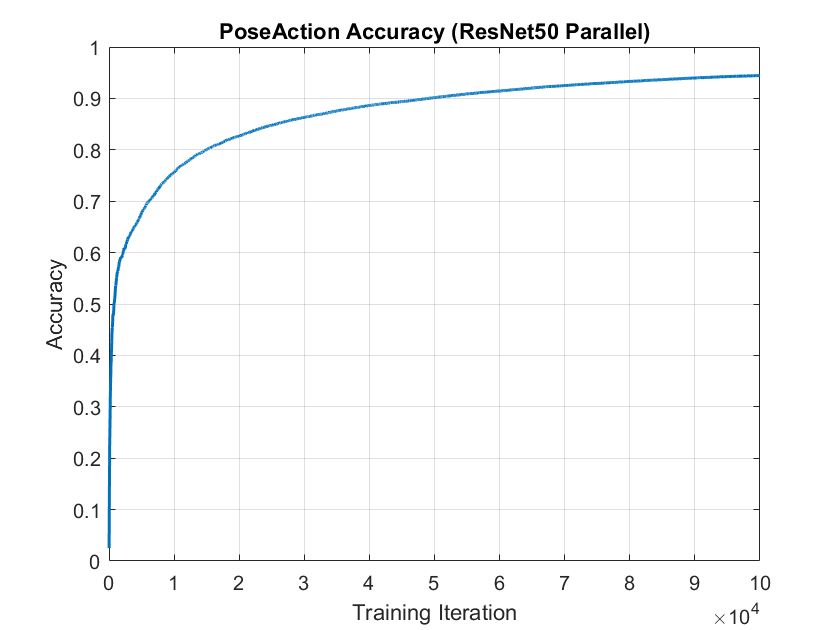}}
\vspace{\BeforeCaptionVSpace}
\caption{PoseAction training accuracy using the ResNet-50 Parallel configuration.} 
\label{fig:results-9} 
\end{figure}

\begin{figure}[H]     
\centerline{\includegraphics[width=0.8\columnwidth]{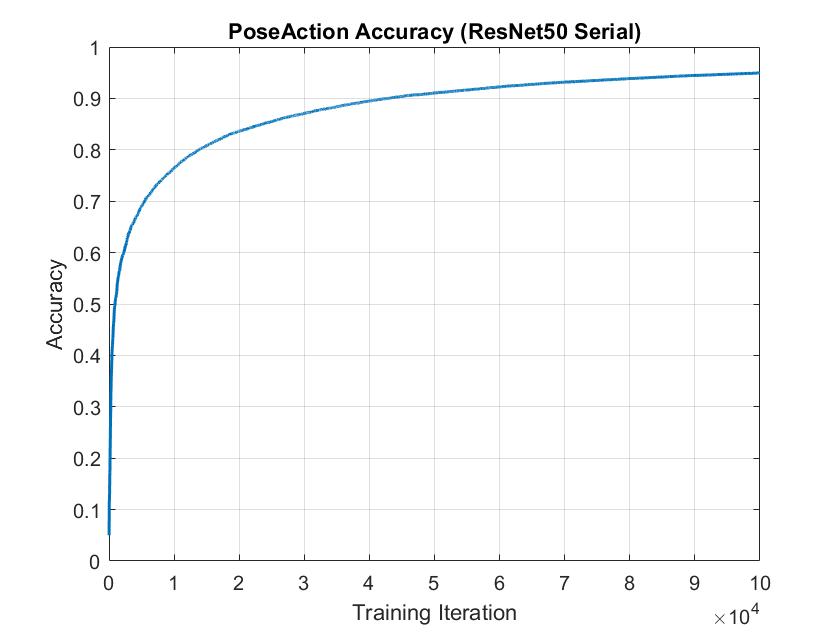}}
\vspace{\BeforeCaptionVSpace}
\caption{PoseAction training accuracy using the ResNet-50 Serial configuration.} 
\label{fig:results-10} 
\end{figure}

\begin{figure}[H]     
\centerline{\includegraphics[width=0.8\columnwidth]{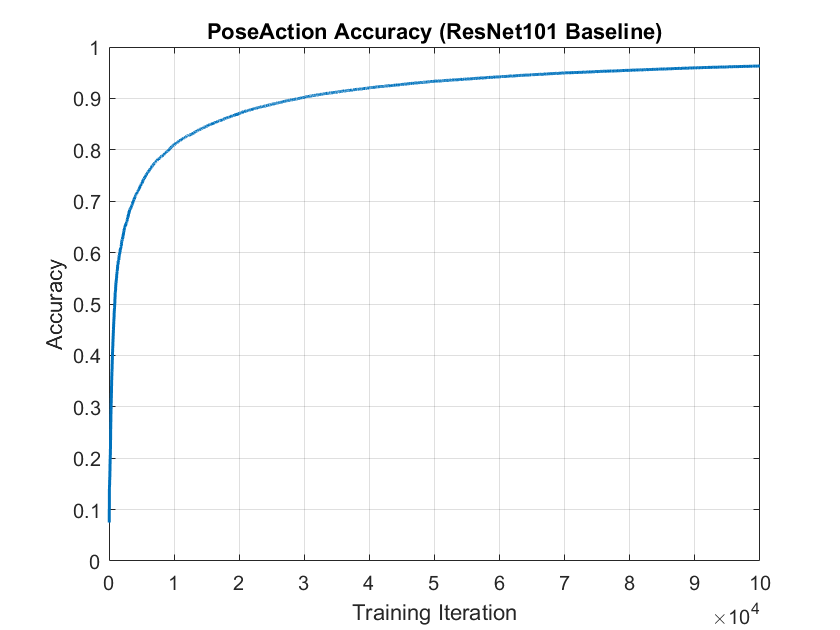}}
\vspace{\BeforeCaptionVSpace}
\caption{PoseAction training accuracy using the ResNet-101 Baseline configuration.} 
\label{fig:results-11} 
\end{figure}

\begin{figure}[H]     
\centerline{\includegraphics[width=0.8\columnwidth]{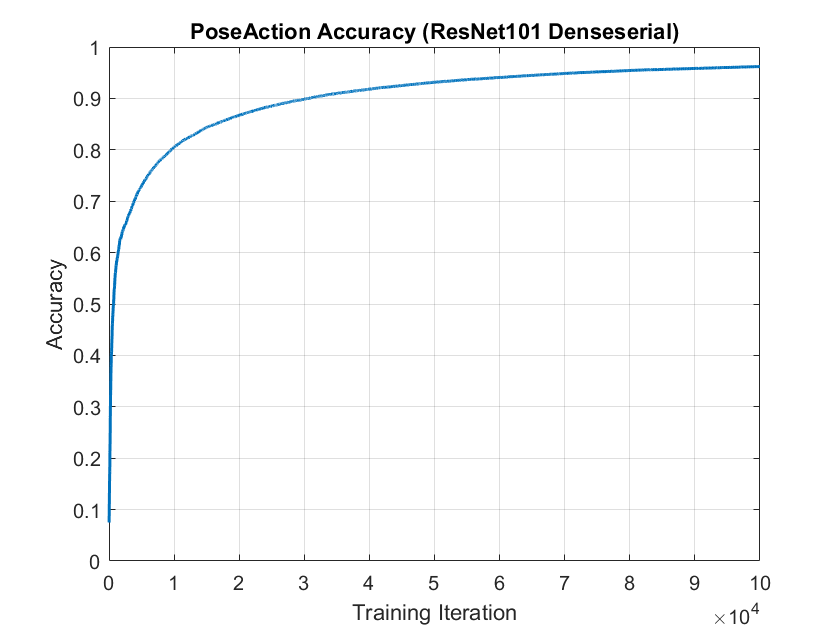}}
\vspace{\BeforeCaptionVSpace}
\caption{PoseAction training accuracy using the ResNet-101 DenseSerial configuration.} 
\label{fig:results-12} 
\end{figure}

\section{Inference Accuracy During Training}
\label{append:infAcc}

\begin{figure}[H]     
\centerline{\includegraphics[width=0.8\columnwidth]{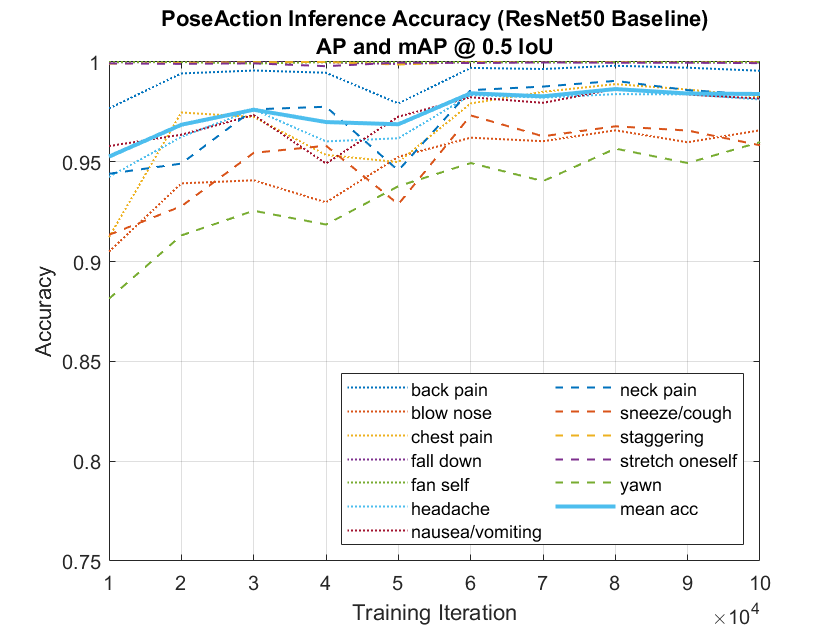}}
\vspace{\BeforeCaptionVSpace}
\caption{PoseAction inference accuracy during training using the ResNet-50 Baseline configuration.} 
\label{fig:results-13} 
\end{figure}

\begin{figure}[H]     
\centerline{\includegraphics[width=0.8\columnwidth]{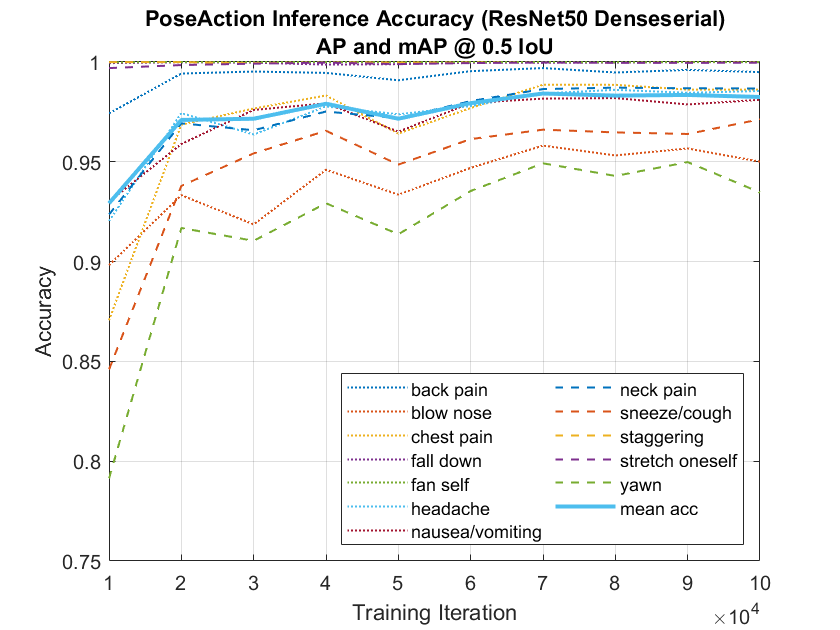}}
\vspace{\BeforeCaptionVSpace}
\caption{PoseAction inference accuracy during training using the ResNet-50 DenseSerial configuration.} 
\label{fig:results-14} 
\end{figure}

\begin{figure}[H]     
\centerline{\includegraphics[width=0.8\columnwidth]{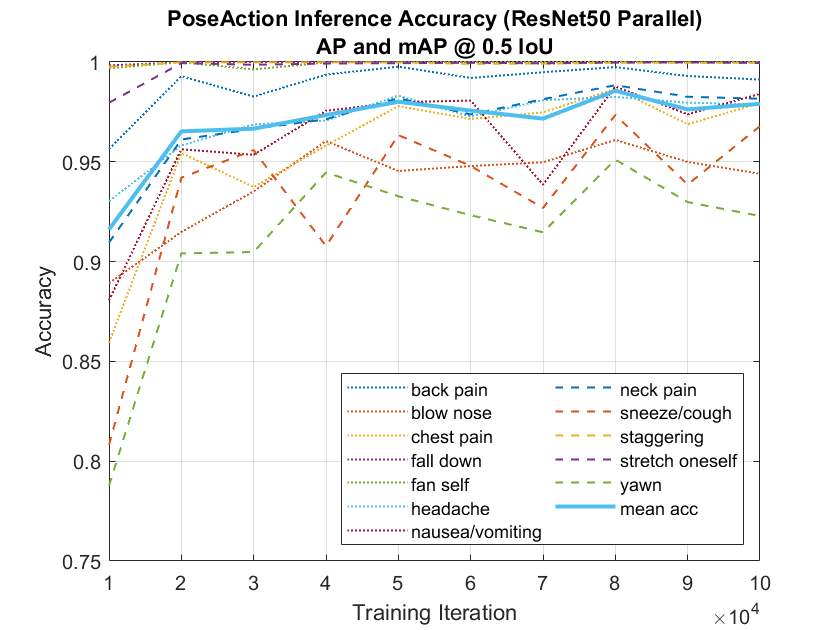}}
\vspace{\BeforeCaptionVSpace}
\caption{PoseAction inference accuracy during training using the ResNet-50 Parallel configuration.} 
\label{fig:results-15} 
\end{figure}

\begin{figure}[H]     
\centerline{\includegraphics[width=0.8\columnwidth]{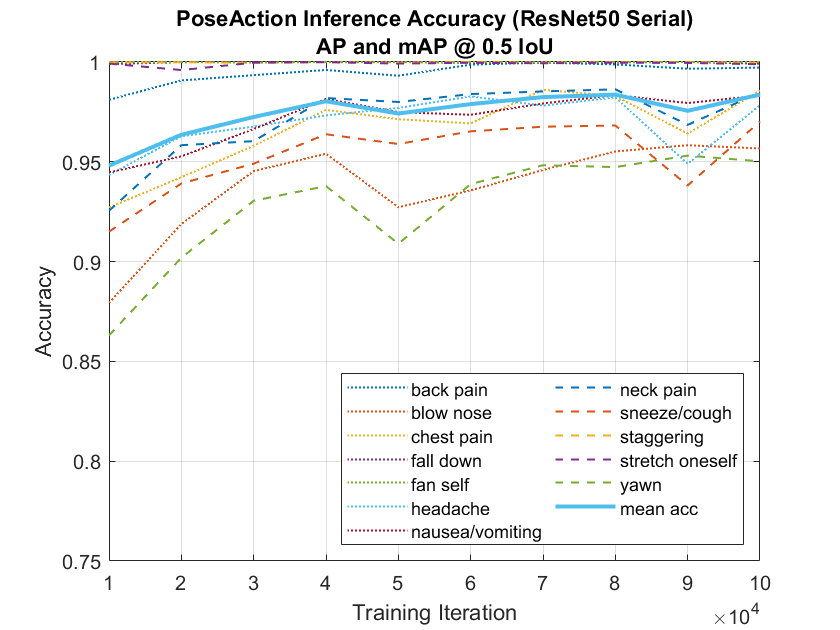}}
\vspace{\BeforeCaptionVSpace}
\caption{PoseAction inference accuracy during training using the ResNet-50 Serial configuration.} 
\label{fig:results-16} 
\end{figure}

\begin{figure}[H]     
\centerline{\includegraphics[width=0.8\columnwidth]{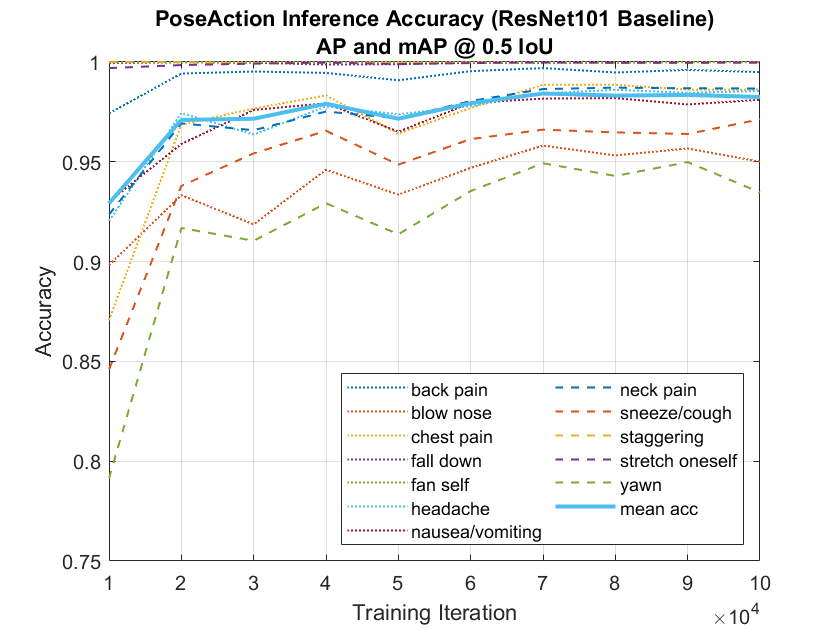}}
\vspace{\BeforeCaptionVSpace}
\caption{PoseAction inference accuracy during training using the ResNet-101 Baseline configuration.} 
\label{fig:results-17} 
\end{figure}

\begin{figure}[H]     
\centerline{\includegraphics[width=0.8\columnwidth]{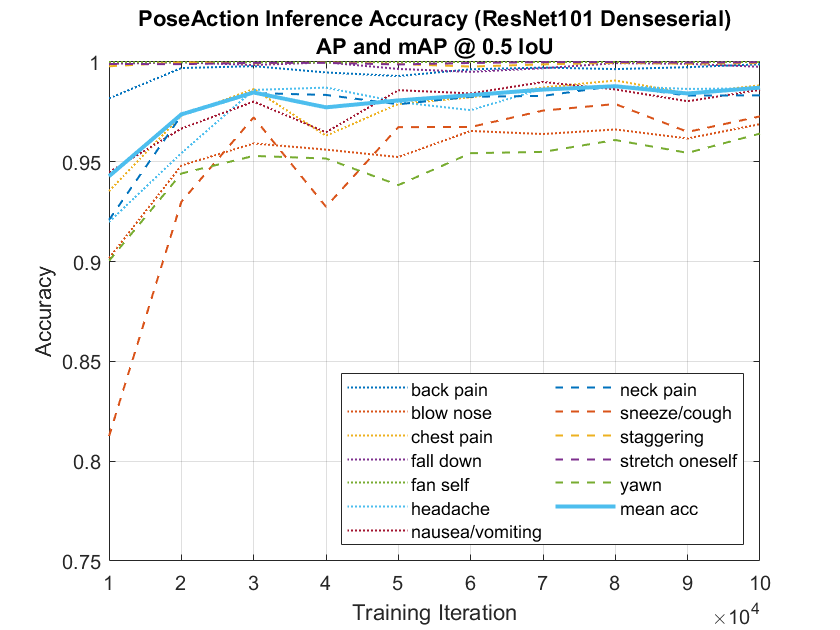}}
\vspace{\BeforeCaptionVSpace}
\caption{PoseAction inference accuracy during training using the ResNet-101 DenseSerial configuration.} 
\label{fig:results-18}
\end{figure}

\section{Confusion Matrices Obtained from Inference with Final Model Parameters}
\label{append:infCM}

\begin{figure}[H]     
\centerline{\includegraphics[width=0.8\columnwidth]{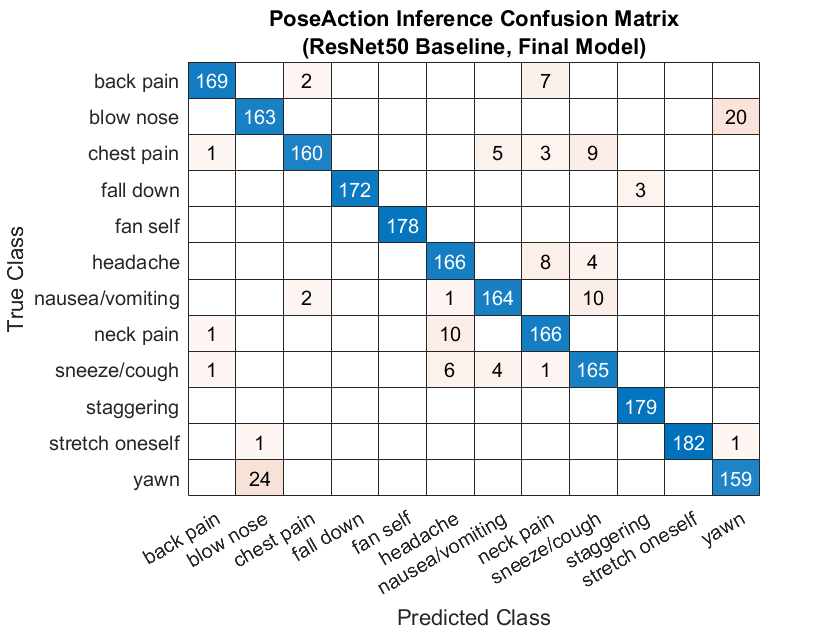}}
\vspace{\BeforeCaptionVSpace}
\caption{Confusion matrices with using the ResNet-50 Baseline configuration.} 
\label{fig:results-19} 
\end{figure}

\begin{figure}[H]     
\centerline{\includegraphics[width=0.8\columnwidth]{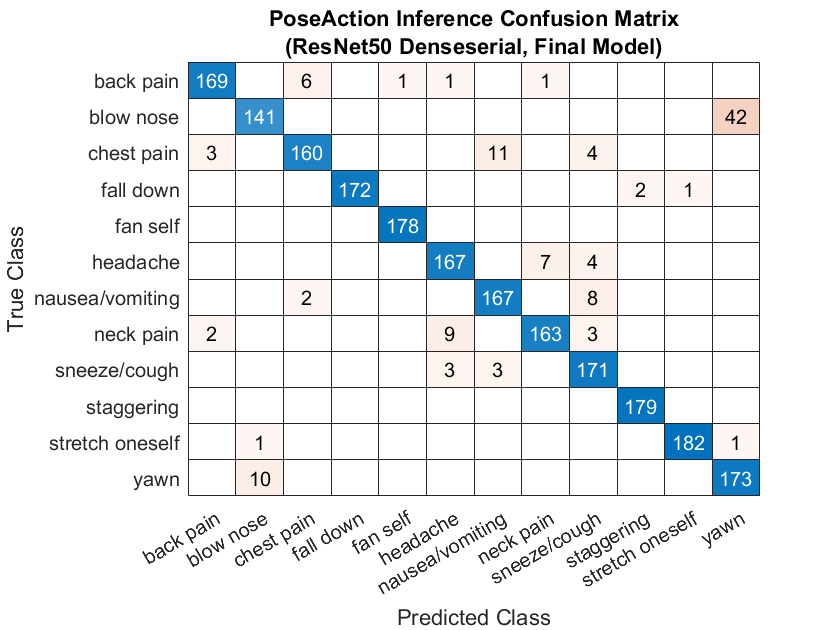}}
\vspace{\BeforeCaptionVSpace}
\caption{Confusion matrices with using the ResNet-50 DenseSerial configuration.} 
\label{fig:results-20} 
\end{figure}

\begin{figure}[H]     
\centerline{\includegraphics[width=0.8\columnwidth]{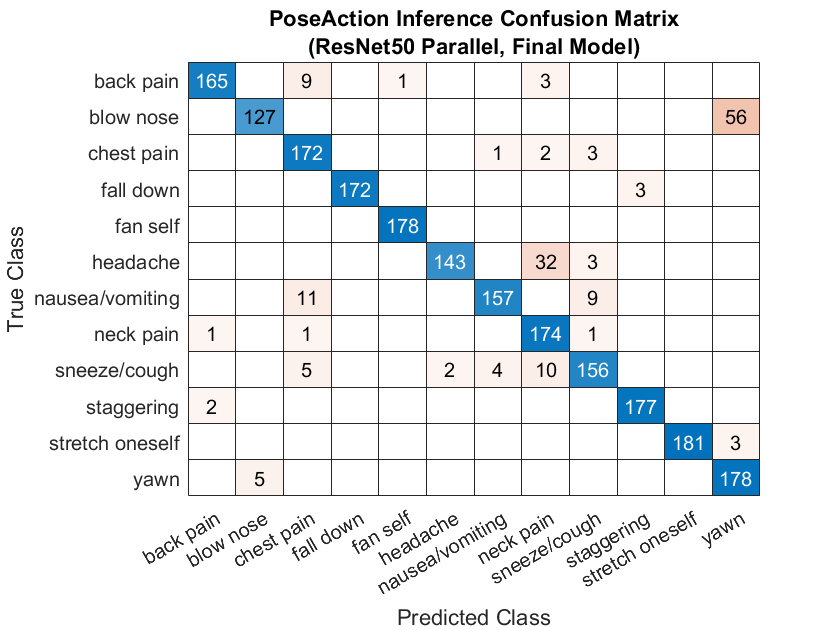}}
\vspace{\BeforeCaptionVSpace}
\caption{Confusion matrices with using the ResNet-50 Parallel configuration.} 
\label{fig:results-21} 
\end{figure}

\begin{figure}[H]     
\centerline{\includegraphics[width=0.8\columnwidth]{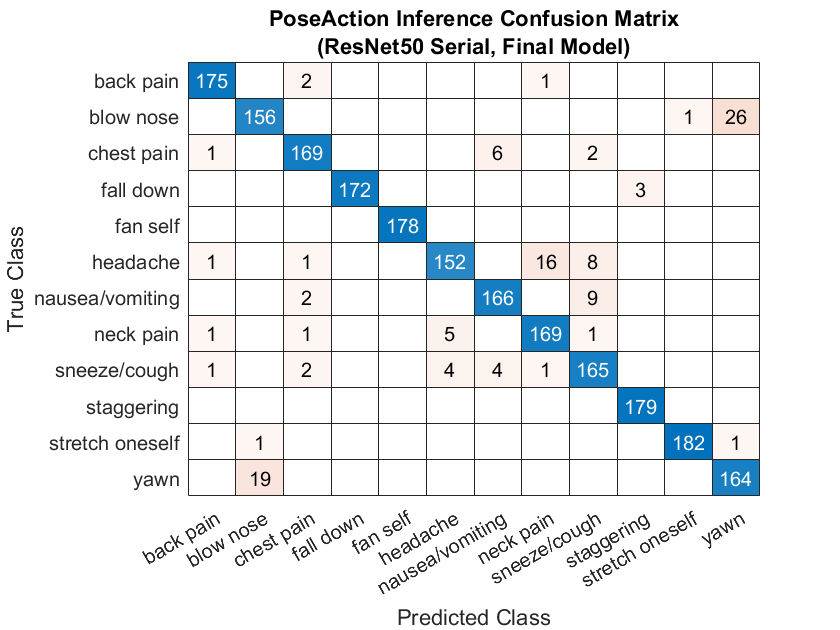}}
\vspace{\BeforeCaptionVSpace}
\caption{Confusion matrices with using the ResNet-50 Serial configuration.} 
\label{fig:results-22} 
\end{figure}

\begin{figure}[H]     
\centerline{\includegraphics[width=0.8\columnwidth]{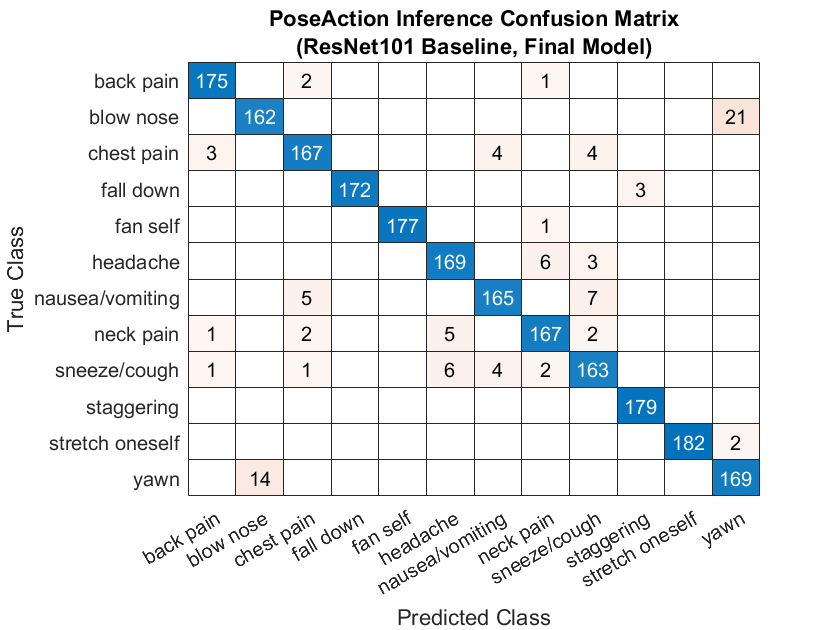}}
\vspace{\BeforeCaptionVSpace}
\caption{Confusion matrices with using the ResNet-101 Baseline configuration.} 
\label{fig:results-23} 
\end{figure}

\begin{figure}[H]     
\centerline{\includegraphics[width=0.8\columnwidth]{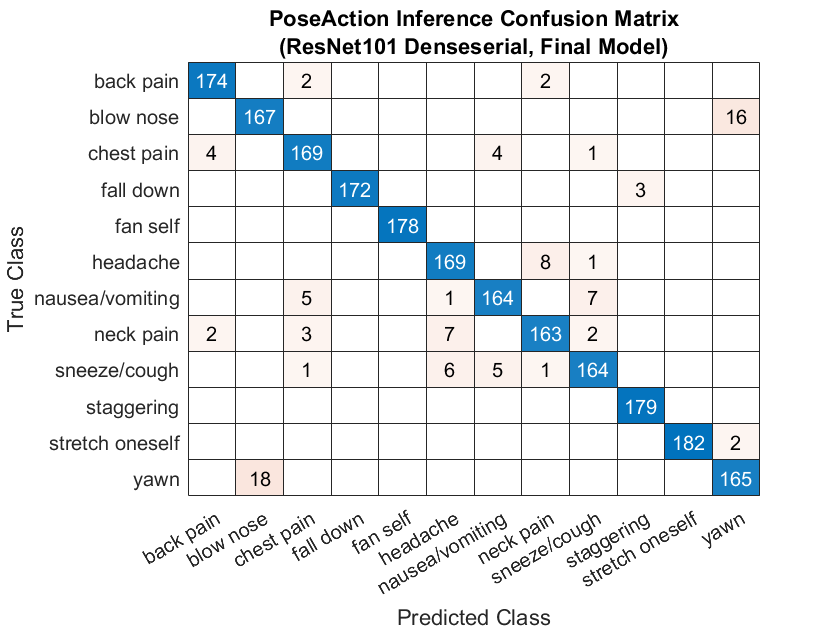}}
\vspace{\BeforeCaptionVSpace}
\caption{Confusion matrices with using the ResNet-101 DenseSerial configuration.} 
\label{fig:results-24} 
\end{figure}

\end{appendices}

\end{document}